%% file: main.tex
\documentclass[10pt,twocolumn,letterpaper]{article}

\usepackage[pagenumbers]{cvpr} 

\input{preamble}

\definecolor{cvprblue}{rgb}{0.21,0.49,0.74}
\usepackage[pagebackref,breaklinks,colorlinks]{hyperref}

\usepackage[capitalize]{cleveref}
\crefname{section}{Sec.}{Secs.}
\Crefname{section}{Section}{Sections}
\Crefname{table}{Table}{Tables}
\crefname{table}{Tab.}{Tabs.}

\def\paperID{*****} 
\def\confName{CVPR}
\def\confYear{2024}

\title{Perceptual Quality Improvement in Videoconferencing \\ using Keyframes-based GAN}

\author{Lorenzo Agnolucci \and Leonardo Galteri \and  Marco Bertini \and Alberto Del Bimbo \vspace{0.1ex} \and
University of Florence - Media Integration and Communication Center (MICC) \\
Florence, Italy\\
{\tt\small [name.surname]@unifi.it}
}

\begin{document}
\maketitle
\input{sec/0_abstract}    
\input{sec/1_intro}
\input{sec/2_related}
\input{sec/3_method}
\input{sec/4_results}
\input{sec/5_conclusion}
{
    \small
    \bibliographystyle{ieeenat_fullname}
    \bibliography{main}
}

\input{sec/X_suppl}

\end{document}

%% file: preamble.tex
\usepackage[dvipsnames]{xcolor}

\usepackage{graphicx,scalerel}
\usepackage{amsmath}
\usepackage{amssymb}
\usepackage{booktabs}

\newcommand{\hide}[1]{}

\usepackage{color}
\usepackage{xcolor,colortbl}
\usepackage{mathtools}
\usepackage{multirow}
\usepackage{bm} 
\definecolor{purple}{rgb}{0.65,0,0.65}
\definecolor{dark_green}{rgb}{0, 0.5, 0}
\definecolor{blueish}{rgb}{0.0, 0.3, .6}
\definecolor{tabhighlight}{HTML}{e5e5e5}
\newcolumntype{h}{>{\columncolor{tabhighlight}}c}

\makeatletter
\newcommand*\wt[2][0.15ex]{%
        \begingroup
        \mathchoice{\wt@helper{#1}{#2}{\displaystyle}{\textfont}}
                   {\wt@helper{#1}{#2}{\textstyle}{\textfont}}
                   {\wt@helper{#1}{#2}{\scriptstyle}{\scriptfont}}
                   {\wt@helper{#1}{#2}{\scriptscriptstyle}{\scriptscriptfont}}%
        \endgroup
        #2%
}
\newcommand*\wt@helper[4]{%
        \def\currentfont{\the#41}%
        \def\currentskewchar{\char\the\skewchar\currentfont}%
        \setbox\tw@\hbox{\currentfont#2\currentskewchar}%
        \dimen@ii\wd\tw@
        \setbox\tw@\hbox{\currentfont#2{}\currentskewchar}%
        \advance\dimen@ii-\wd\tw@
        \rlap{\raisebox{-#1}{$\m@th#3\kern\dimen@ii\widetilde{\phantom{#2}}$}}%
}
\makeatother

\usepackage[export]{adjustbox} 

\newlength{\gridimagewidth}
\usepackage{array}
\newcolumntype{C}[1]{>{\centering\let\newline\\\arraybackslash\hspace{0pt}}m{#1}}

\newcolumntype{h}{>{\columncolor{tabhighlight}}c}

%% file: sec/0_abstract.tex
\begin{abstract}
In the latest years, videoconferencing has taken a fundamental role in interpersonal relations, both for personal and business purposes. Lossy video compression algorithms are the enabling technology for videoconferencing, as they reduce the bandwidth required for real-time video streaming. However, lossy video compression decreases the perceived visual quality. Thus, many techniques for reducing compression artifacts and improving video visual quality have been proposed in recent years. In this work, we propose a novel GAN-based method for compression artifacts reduction in videoconferencing. Given that, in this context, the speaker is typically in front of the camera and remains the same for the entire duration of the transmission, we can maintain a set of reference keyframes of the person from the higher-quality I-frames that are transmitted within the video stream and exploit them to guide the visual quality improvement; a novel aspect of this approach is the update policy that maintains and updates a compact and effective set of reference keyframes.
First, we extract multi-scale features from the compressed and reference frames. Then, our architecture combines these features in a progressive manner according to facial landmarks. This allows the restoration of the high-frequency details lost after the video compression.
Experiments show that the proposed approach improves visual quality and generates photo-realistic results even with high compression rates. Code and pre-trained networks are publicly available at \small{\href{https://github.com/LorenzoAgnolucci/Keyframes-GAN}{\url{https://github.com/LorenzoAgnolucci/Keyframes-GAN}}}.
\end{abstract}

%% file: sec/1_intro.tex
\section{Introduction} \label{sec:intro}
In recent years, videoconferencing has become a primary means of personal and business communication all over the world, also because of the emergence of the COVID-19 pandemic.

Lossy video compression algorithms such as H.264 and H.265 allow for a decrease in the bandwidth required for video transmission but introduce compression artifacts that reduce the perceived quality of the video stream. The degradation of the visual quality worsens the user experience, even making it unacceptable in certain cases.

For these reasons, the development of methods for video quality enhancement constitutes a very active area of research. In recent years, Generative Adversarial Networks (GANs) have emerged as one of the most promising and powerful tools for several image and video processing tasks, thanks to their ability to generate photorealistic and perceptually satisfying results \cite{galteri2017deep, galteri2019deep, liu2020generative}.

Applying deep learning-based enhancement methods to videos has several advantages. Firstly, these methods can be applied as post-processing steps to existing video compression and transmission systems without requiring to change any component and being independent of the specific video codec employed. Secondly, enhancing the visual quality of videos reduces compression artifacts and other types of degradation, thus improving the user experience. Finally, the improvement in the perceived quality makes it possible to transmit videos with higher compression rates, consequently reducing the needed bandwidth. For example, \cite{galteri2020increasing} uses semantic video coding and a GAN to obtain a  quality comparable to the one obtained by standard H.264 with three times the bandwidth. \cite{wang2021oneshot} proposes a talking-head synthesis approach that reconstructs a video using one-tenth of the original bandwidth.

\input{fig/teaser}

\paragraph*{Contributions}
In this work, we propose a novel GAN-based approach for improving visual quality in videoconferencing. In videoconferencing the background has so little relevance \cite{wijnants2019talking} that some commercial solutions provide features to blur or replace the background with a virtual one. 
For this reason, we focus on the enhancement of the framed person, and in particular on the head area, because it is the most expressive and important part of interpersonal communications.
Our approach is based on the assumption that the subject speaking in front of the camera stays the same for a relatively long consecutive time frame so that we can exploit for enhancement the previous high-quality reference keyframes of the Group of Pictures (GOP) coding (\ie the so-called \textit{I-frames}), used in video compression algorithms as the base for motion-based compression.  In particular, we propose a novel policy to create and update a set of reference keyframes in order to keep this set small, and thus memory efficient, and also to make it effective for the improvement of the visual quality. Our model extracts multi-scale features of the compressed frame and a reference keyframe and then combines them according to the facial landmarks (see \cref{fig:teaser}). The feature fusion is performed with Adaptive Spatial Feature Fusion (ASFF) \cite{li2020enhanced} and Spatial Feature Transform (SFT) \cite{wang2018recovering} blocks in a progressive manner that helps in restoring coarse-to-fine details. We designed a pipeline for video enhancement that involves preserving a limited number of keyframes extracted from the video stream and using the most useful ones as a reference for restoring the compressed frame. The experiments and the comparison with competing state-of-the-art approaches show that our proposed method is very effective in generating photo-realistic results even with high compression rates. 

%% file: fig/teaser.tex
\begin{figure*}[t]
\centering
\includegraphics[width=0.99\textwidth]{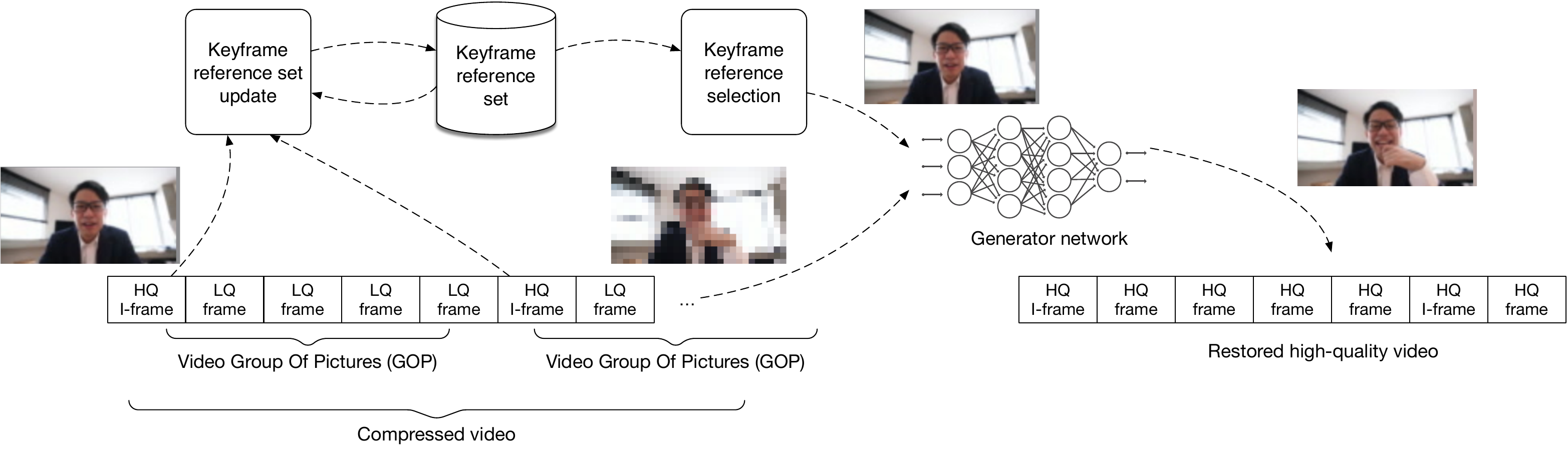}
\caption{Overview of the proposed system at runtime. High quality reference keyframes (video I-frames) are used in our GAN-based approach to improve the visual quality of the video conference stream. The algorithm used to update the keyframe reference set is a key element to improve the visual quality of the restored frames.}
\label{fig:teaser}
\end{figure*}

%% file: sec/2_related.tex
\section{Related Work}\label{sec:related}

\paragraph{Video Coding}
Some interesting initial works have addressed the quality improvement of videos and images using coding based on neural networks \cite{rippel2017real, rippel2019learned}. These approaches are currently not deployable with satisfying visual results due to an unbearable computational cost. Moreover, fully learned compression requires the standardization and diffusion of a novel technology, which is a very high market barrier to practical use. 

\paragraph{Video Quality Improvement}
Recently,  many learning-based image enhancement techniques have been proposed \cite{kang-2015, dong2015compression, MaoSY16a, svoboda2016compression, wang2016d3, galteri2017deep, yu2017convolutional, galteri2019deep, cavigelli2017cas, yoo2018image, maleki2018blockcnn, vaccaro-2021}. Such approaches learn deep convolutional architectures, often based on GANs, to restore low-quality images corrupted by compression artifacts into high-quality ones, and deal with generic video content.
\cite{guan2019mfqe} presents a Multi-Frame Quality Enhancement approach for compressed videos. After observing that the quality of compressed videos fluctuates across frames, the authors developed a BiLSTM-based detector to locate Peak Quality Frames (PQFs), that is frames that have a higher quality than their neighbors, whose information can be exploited to reduce the distortion of low-quality frames. A non-PQF and its nearest two PQFs are the input of a multi-frame CNN, composed of a motion compensation and a quality enhancement subnet. \cite{wang2019edvr} presents EDVR, a video restoration framework with enhanced deformable convolutions. A pyramid, cascading and deformable module uses deformable convolutions in a coarse-to-fine manner to align the features of the reference frame to that of its neighboring frames and then a temporal and spatial attention fusion module combines them.

\paragraph{Face Quality Improvement}
Face super-resolution has been addressed in \cite{dogan2019exemplar}, where the authors proposed GWAInet, a GAN-based approach that performs $8\times$ face super-resolution using a HR reference image of the same person depicted in the LR image. A warper subnetwork aligns the contents of the reference image to the input image. Then, after extracting the features of the LR and HR images, a feature fusion chain combines them to exploit the reference image. A peculiarity of this method is that it does not require facial landmarks for the training. 
In \cite{li2019recovering} super-resolution of extremely degraded faces is dealt with a GAN that produces a coarse SR image. Then, the result is refined by exploiting facial components extracted from multiple high-quality warped images of the same person or a similar one.
In \cite{yang2020hifacegan} the problem of face quality improvement is formulated as a dual-blind restoration problem, lifting the requirements of both the degradation and structural prior for training. The authors present HiFaceGAN, a collaborative suppression and replenishment framework with a nested architecture for multi-stage face renovation with hierarchical semantic guidance.
\cite{yang2021GPEN} proposes a GAN prior embedded network for blind face restoration, using a U-shaped DNN for face restoration as a decoder. 
PSFR-GAN, a GAN-based Progressive Semantic-aware Style Transformation framework presented in \cite{chen2020progressive}, uses a face parsing network to obtain a segmentation map given an LQ face image. The input image and the segmentation map are exploited to produce a multi-scale pyramid of the inputs modulating different scale features with a semantic-aware style transfer approach. A semantic aware style loss accounts for each semantic region individually.
In \cite{li2018learning} blind face restoration task is tackled with a Guided Face Restoration Network (GFRNet) that takes advantage of a high-quality reference image of the same identity. A warper subnetwork reduces the difference in pose and expression between the two images to better recover fine and identity-aware facial details with a reconstruction subnetwork.
The Deep Face Dictionary Network (DFDNet) proposed in \cite{li2020blind} attempts to overcome the main limitation of reference-based methods by observing that facial components are similar between different people. Multi-scale dictionaries of facial parts are built offline with K-means from high-quality images. The features in the dictionaries most similar to the facial components of the degraded input are leveraged for restoration by means of Dictionary Feature Transfer and Spatial Feature Transform blocks. 
In \cite{li2020enhanced} blind face restoration is tackled by exploiting a high-quality image selected from multiple available images of the same person as a reference to restore a degraded one. The features of the guidance image are warped to the low-quality ones according to the facial landmarks to reduce the difference in pose and expression. Multiple Adaptive Spatial Feature Fusion blocks combine the degraded and guidance features by generating an attention mask with facial landmarks to guide the restoration of the facial components.
In \cite{galteri2020increasing} a method that combines semantic video coding and GAN-based video quality restoration is proposed for video conference systems, using a perceptual loss that accounts separately for the background and the foreground face.
\cite{doukas2020headgan} presents HeadGAN, a method for head reenactment that conditions head synthesis on 3D face representations from a driving video. Audio features are exploited to better synthesize mouth movements. When driving and reference identities coincide, HeadGAN can be used for face reconstruction.
In \cite{wang2021gfpgan} facial priors encapsulated in a pre-trained GAN (GFP-GAN) are incorporated for blind face restoration by means of channel-split Spatial Feature Transform layers. Unlike GAN inversion methods, GFP-GAN can restore faces with a single forward pass.
\cite{liu2021face} tackles blind face restoration with a GAN that uses multi-scale facial features. A feature prior loss aims to reduce the difference in the feature space between the input and restored images, thus preserving the overall image content and spatial structure information. 
\cite{li2021universal} proposes a restoration with a memorized modulation framework for blind face restoration. Low-level spatial feature embedding, wavelet memory embedding, and disentangled high-level noise embedding are combined with adaptive attention maps.
\cite{zhang2020davdnet} presents DAVD-Net, a DCNN architecture that exploits the audio-video correlations to remove compression artifacts in close-up talking head videos. The audio features are extracted with a BiLSTM and organized in a 2D form. The video and audio features are aggregated with a spatial attention module. To further improve the restoration the structural information of the encoder in the video compression standards is embedded into the network by adding a constraining projection module. 
In \cite{liu2021mrs} face quality of compressed videos is enhanced with MRS-Net+, a multi-level architecture comprised of one base and two refined enhancement levels that restore small, medium, and large-scale faces, respectively. A landmark-assisted pyramid alignment subnet is developed to align faces across consecutive frames. 
\cite{zhang2021multi} and \cite{guo2020deep} exploit a multi-modality neural network to restore strongly compressed face videos. They both use video and audio signals, combined with codec information in \cite{zhang2021multi} and with an emotion state in \cite{guo2020deep}.
\cite{yasarla2021network} presents a multi-task face restoration network that relies on network architecture search to restore images affected by various degradations. Additionally, during training clean images of the same subject as the degraded image are exploited by means of an identity loss.
\cite{wang2022restoreformer} proposes a method based on fully-spatial attention to tackle blind face restoration. A multi-head cross-attention layer takes the features of a degraded face as queries while the key-value pairs are from high-quality facial priors. The key-value pairs are sampled from a reconstruction-oriented high-quality dictionary.

\begin{figure*}[!htb]
    \centering
    \makebox[0.96\columnwidth][c]{\includegraphics[width=\textwidth]{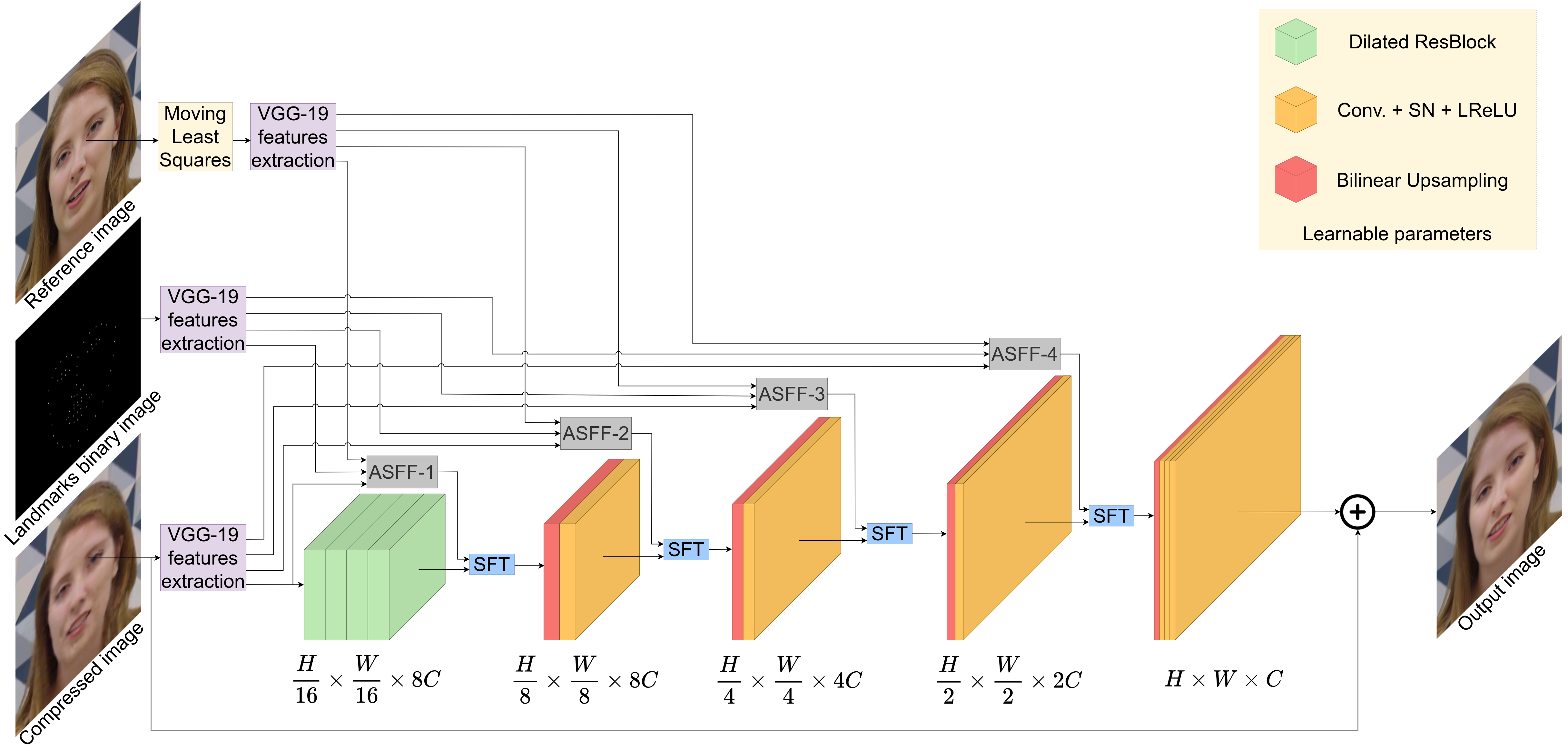}}\vspace{-8pt}
    \caption{Overview of the proposed architecture. Best viewed in color on PDF.}
    \label{fig:architecture}\vspace{-6pt}
\end{figure*}

\smallskip
Even if our aim is to improve the perceptual quality of videos we did not follow the standard multi-frame restoration approach that is commonly used in video restoration tasks, such as in MFQE 2.0 \cite{guan2019mfqe}, MRS-NET+ \cite{liu2021mrs} or DAVD-Net \cite{zhang2020davdnet}, because it usually involves looking also at future frames and this is not possible in a real-time stream. Surely taking into account only past neighboring frames is a possibility, but we preferred to consider possibly very distant I-frames and not necessarily the closest one. This preference is possible in videoconferencing because the subject usually is the same for the entire transmission so old I-frames can still be very useful in restoring the current compressed frame. This is similar to exemplar-guided face image restoration techniques but given that our method is applied to videos we can exploit multiple I-frames from the same video stream as possible references, dynamically updating the set of keyframes with the policy we designed to obtain the best performance. Precisely the LFU-inspired update strategy for the dynamic set of keyframes is what mainly differentiates our work from exemplar-based face restoration methods that constitute the state-of-the-art. For instance, ASFFNet \cite{li2020enhanced} relies on a given set of reference images representing the same person and it can not handle a dynamic set of references, nor a policy for updating it. Similarly, DFDNet \cite{li2020blind} needs an offline-generated dictionary of features of different subjects, therefore it can not exploit high-quality I-frames of the same subject that arrive in real-time.

%% file: sec/3_method.tex
\section{Proposed Approach}\label{sec:method}

Since its introduction in \cite{goodfellow2014generative}, the Generative Adversarial Network (GAN) framework has emerged as a powerful tool for various image and video synthesis tasks, such as image-to-image translation \cite{isola2017image}, face reenactment \cite{wu2018reenactgan} and pose transfer \cite{wang2019fewshot}. Compared to other deep generative models, like Deep Boltzmann Machines \cite{fischer2012introduction} or Variational AutoEncoders \cite{kingma2019introduction}, GANs proved to be able to generate more photorealistic results \cite{goodfellow2016nips, liu2020generative}, and have been successfully used to improve the visual quality of images \cite{galteri2019deep} and videos \cite{vaccaro-2021}.
Our method is based on such a framework.

\subsection{Proposed Architecture}\label{sec_architecture}
We propose a novel GAN architecture shown in \cref{fig:architecture} and inspired by \cite{li2020enhanced} and \cite{li2020blind}. 
Similarly to \cite{li2020enhanced}, we adopt the ASFF block and Moving Least Squares for warping. Differently from \cite{li2020enhanced}, we warp directly the reference image and not its features and we extract and fuse features at multiple scales in a progressive manner to help the network in restoring coarse-to-fine details.
We took inspiration from \cite{li2020blind} in the use of multi-scale features and of the SFT block, but we leverage a high-quality image of the same person to better restore subject-specific details. Differently from both \cite{li2020enhanced} and \cite{li2020blind}, we select our reference image from the best-performing set of high-quality keyframes coming from the same video, which is built and updated with our proposed policy. 

Our architecture is based on U-Net \cite{ronneberger2015u} and it is composed of an encoder, that processes the input so that it is smaller in terms of spatial dimensions but deeper in terms of the number of channels, and by a decoder, that inverts the process. Multi-scale reference features are combined with the features of the degraded image in a progressive manner. This approach can make the network learn coarse-to-fine details and is beneficial to the restoration process. \\
Our model takes 3 inputs:
\begin{itemize}
    \item a degraded (\ie highly compressed) image;
    \item a high-quality reference image (\ie a video \textit{I-frame});
    \item a binary image that is white only in correspondence with the facial landmarks of the compressed image.
\end{itemize}
The model produces a restored image from the compressed one.

\begin{figure}
    \centering
    \includegraphics[width=\columnwidth]{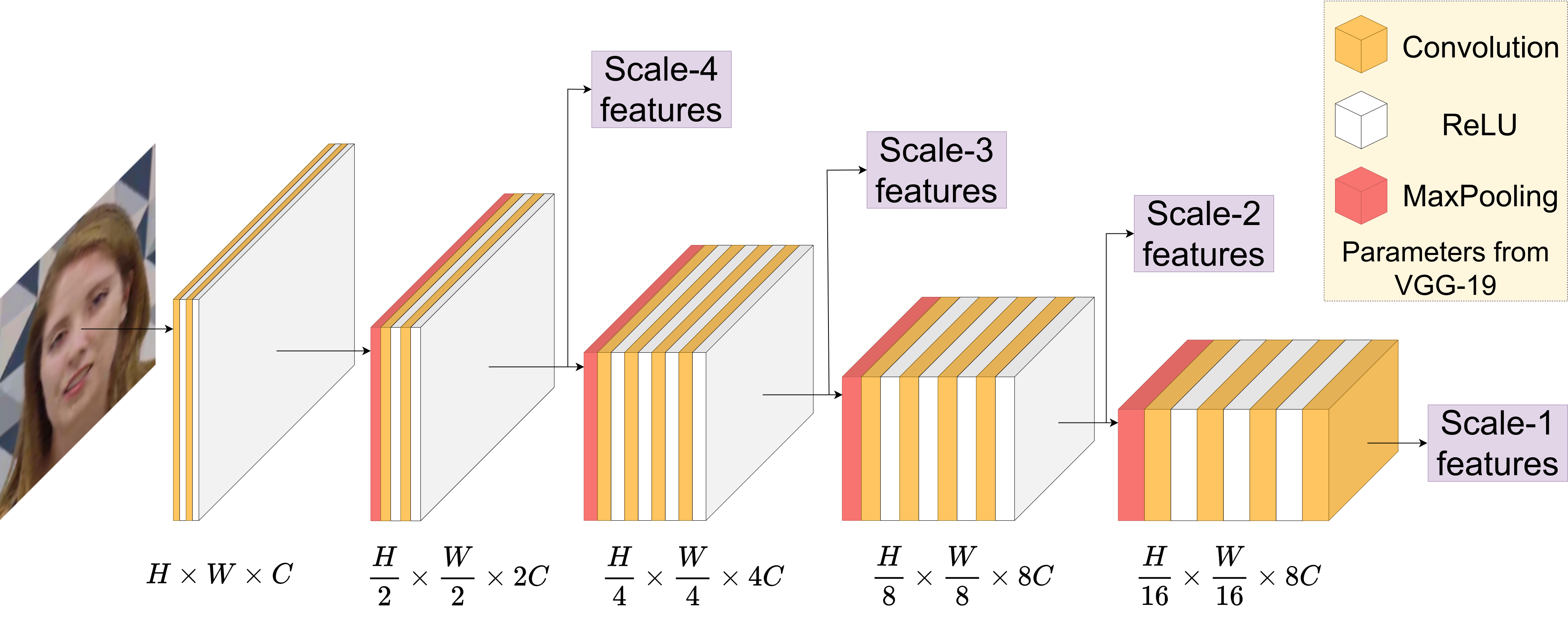}
    \caption{Diagram of the multi-scale feature extraction with VGG-19}
    \label{fig:VGG_features_extraction}\vspace{-6pt}
\end{figure}

We use a pre-trained VGG-19 \cite{simonyan2014very} to extract multi-scale features from the degraded, reference and landmarks binary images. The reference (guidance) image is previously warped to the degraded one based on the facial landmarks using Moving Least Squares (\cref{sec:MLS}). We extract features at 4 different scales from the layers \texttt{relu$\_$2$\_$2}, \texttt{relu$\_$3$\_$4}, \texttt{relu$\_$4$\_$4} and \texttt{conv$\_$5$\_$4} of the VGG-19. The feature extraction is depicted in \cref{fig:VGG_features_extraction}.

To align the warped reference and degraded features we adopt AdaIN \cite{huang2017arbitrary}. This helps reduce the difference in style and illumination between the two images and thus improves the restoration. We denote by $F^{d}$ and $F^{g}$ the degraded and guidance features. The AdaIN can be written as
\begin{equation}
    F^{g, a} = \sigma(F^{d}) \left ( \frac{F^{g}- \mu(F^{g})}{\sigma(F^{g})} \right ) + \mu(F^{d})
\end{equation}
where $\sigma(\cdot)$ and $\mu(\cdot)$ represent the mean and the standard deviation.

After going through multiple dilated residual blocks, the degraded features are progressively upsampled by enlarging the spatial resolution and reducing the number of channels. At the same time, they are combined with the reference features by means of Adaptive Spatial Feature Fusion (\cref{sec:ASFF}) and Spatial Feature Transform (SFT) \cite{wang2018recovering} blocks.

\begin{figure}
    \centering
    \includegraphics[width=0.8\columnwidth]{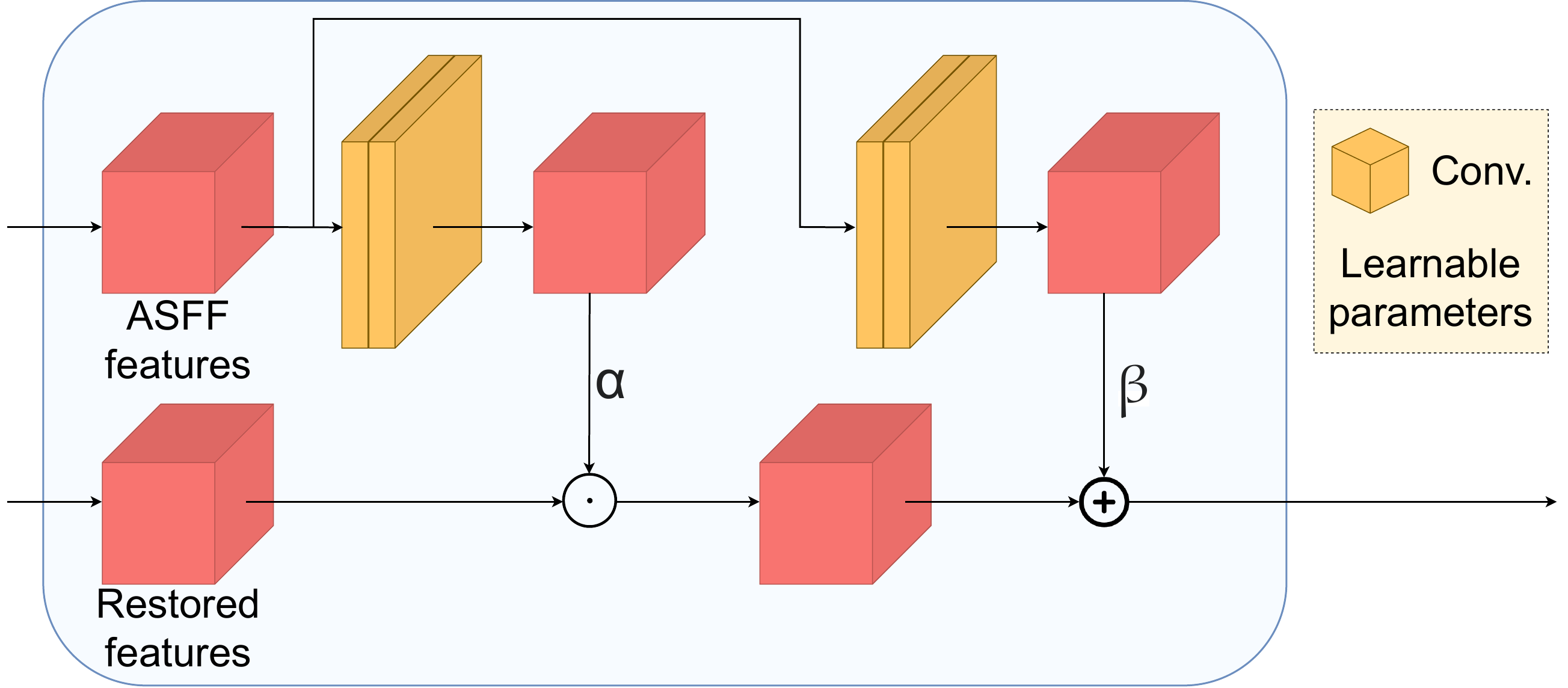}
    \caption{Structure of the SFT block}
    \label{fig:SFT_block}
\end{figure}

The SFT block generates affine transformation parameters for spatial-wise feature modulation incorporating some prior condition. The scale $\alpha$ and the shift $\beta$ parameters are learned from the features outputted by the corresponding ASFF block. The output of the SFT block is formulated as
\begin{equation}
    SFT = \alpha \odot F^{r} + \beta
\end{equation}
where $\odot$ is the element-wise product and $F^{r}$ are the restored features, that is the features originated from the degraded ones and restored in the decoding part of the architecture. \Cref{fig:SFT_block} shows the structure of the SFT block.

Following \cite{galteri2020increasing}, we train the network to learn the residual image, so there is a skip connection between the degraded image and the restored output. This choice reduces the overall training time and improves its stability.

\subsection{ASFF Block}
\label{sec:ASFF}
The fusion of the features of the reference and degraded images is a fundamental part of exemplar-based approaches, as it allows to fully exploit the information supplied by the guidance image. Adopting a concatenation-based approach, as in \cite{dogan2019exemplar, li2018learning}, does not take full advantage of the reference features.

\begin{figure}
    \centering
    \includegraphics[width=0.9\columnwidth]{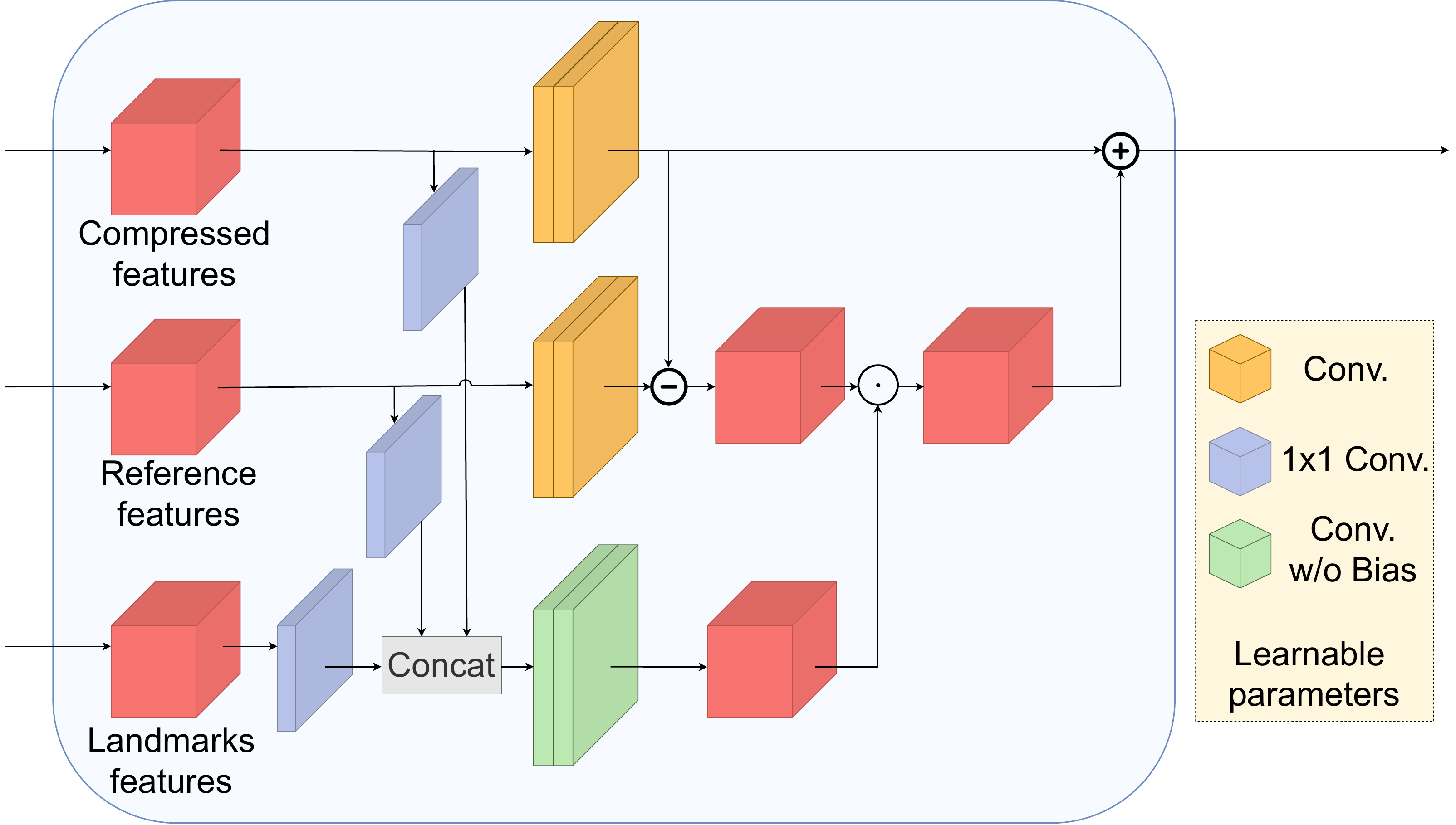}
    \caption{Structure of the ASFF block}
    \label{fig:ASFF_block}
\end{figure}

Thus, in our multi-scale architecture, we rely on multiple Adaptive Spatial Feature Fusion (ASFF) blocks \cite{li2020enhanced}. While the reference image generally contains more high-quality details, the degraded image should have more weight in the reconstruction of the overall face components. For example, if the mouth of the reference image is closed while that of the compressed image is open, the reconstruction of the teeth should be mainly based on the restored features from the degraded image. For this reason, ASFF blocks generate an attention mask based on the degraded image facial landmarks to guide the fusion of the guidance and restored features. 
\Cref{fig:ASFF_block} shows the structure of the ASFF block.

\subsection{Warping Reference with Moving Least Squares}
\label{sec:MLS}

For most guided face restoration methods, the performance is diminished by the pose and expression difference between reference and degraded images because it introduces artifacts in the reconstruction result. Thus, we spatially aligned the reference and compressed images with an image deformation method based on Moving Least Squares (MLS) \cite{schaefer2006image}.

Let $p$ and $q$ be respectively the sets of facial landmarks of the reference and degraded image, with $|p|=|q|=N$. In our case, $N=68$. We aim to find a deformation function $f$ to apply to all the points of the reference image. Given a point $v$ in the image, we solve for the best affine transformation $l_{v}(x)$ that minimizes
\begin{equation} \label{eq:MLS_first_formulation}
    \sum_{i=1}^{N}w_{i} \left| l_{v}(p_{i}) - q_{i} \right| ^{2} \; \text{  where  }\; w_{i} = \frac{1}{\left| p_{i} - v \right| ^ {2}} 
\end{equation}
Because the weights $w_{i}$ are dependent on the point of evaluation $v$ we obtain a different transformation $l_{v}(x)$ for each $v$. We define the deformation function $f$ to be $f(v) = l_{v}(v)$.

Since $l_{v}(x)$ is an affine transformation we can rewrite it in terms of a linear transformation matrix $M$
\begin{equation}
    l_{v}(x) = (x - p_{*}) M + q_{*}
\end{equation}
where $p_{*}$ and $q_{*}$ are weighted centroids
\begin{equation*}
    p_{*} = \frac{\sum_{i=1}^{N}w_{i} p_{i}}{\sum_{i=1}^{N}w_{i}}    \hspace{70pt}
    q_{*} = \frac{\sum_{i=1}^{N}w_{i} q_{i}}{\sum_{i=1}^{N}w_{i}}
\end{equation*}
Based on this insight, the least squares problem of \cref{eq:MLS_first_formulation} can be rewritten as
\begin{equation} \label{eq:MLS_last_formulation}
    \sum_{i=1}^{N}w_{i} \left| \hat{p}_{i} M - \hat{q}_{i} \right| ^{2}
\end{equation}
where $\hat{p}_{i} = p_{i} - p_{*}$ and $\hat{q}_{i} = q_{i} - q_{*}$. The affine deformation that minimizes \cref{eq:MLS_last_formulation} is 
\begin{equation*}
    M = \left ( \sum_{i=1}^{N} \hat{p}_{i}^{T}w_{i}\hat{p}_{i} \right ) ^{-1} \sum_{j=1}^{N}w_{j} \hat{p}_{j}^{T} \hat{q}_{j}
\end{equation*}
With this closed-form solution for M, we can write a simple expression for the deformation function $f$
\begin{equation}
    f(v) = (v - p_{*}) \left ( \sum_{i=1}^{N} \hat{p}_{i}^{T}w_{i}\hat{p}_{i} \right ) ^{-1} \sum_{j=1}^{N}w_{j} \hat{p}_{j}^{T} \hat{q}_{j} + q_{*}
\end{equation}
Applying this deformation function to each point of the reference image lets to warp it according to the facial landmarks of the degraded image.

\subsection{Keyframes Selection and Set Maintenance}\label{sec:keyframes_selection}
Although warping with MLS helps to reduce the distance between the compressed and reference images, if they are too different the results will still be sub-optimal. Thus it is natural to select the optimal reference keyframe as the one that has a similar pose and expression to the degraded image, instead of simply using the previous keyframe. We measure the similarity between a keyframe and the degraded frame with the Euclidean distance between the sets of facial landmarks. Considering videoconferencing, assuming that the talking subject stays the same, even very old keyframes can be useful. So, as the video progresses, one can save a limited set of keyframes, to reduce memory requirements, and then use the most similar one as a reference to restore the current compressed frame. This novel method is the key to improving the overall restoration quality of the video and limits the cases in which the compressed and reference frames are very different.

We took inspiration from the Least-Frequently Used (LFU) cache replacement strategy: for each keyframe of the set, we keep count of how many times it was selected for reconstruction and when a new keyframe is received from the video stream the least used is evicted. However, in this way, the first keyframes of the video would be excessively rewarded. Indeed, since for the first seconds of the video they are the only ones available as a reference they can be used not because of similarity with the compressed frame but for lack of alternatives. To overcome this problem we apply an exponential decay to the number of uses, \ie when a new keyframe arrives the counter of the number of uses of all the keyframes of the set is halved.

\subsection{Training Losses}\label{sec:loss}
As in \cite{li2020enhanced}, to train our model we employed a weighted sum of reconstruction and photo-realistic losses. We denote by $I_{D}$, $I_{R}$ and $I_{GT}$ the degraded, reconstructed and ground-truth (\ie high-quality uncompressed) images, respectively.

The reconstruction loss constrains the reconstructed image to faithfully approximate the ground-truth one and is composed of two terms. First, we relied on the Mean Square Error (MSE), defined as
\begin{equation}
    \ell_{MSE} = \frac{1}{CHW} \left\| I_{R} - I_{GT} \right\| ^{2}
\end{equation}
where $C$, $H$ and $W$ denote the channel, height and width of the image. Second, we adopted the perceptual loss \cite{johnson2016perceptual, ledig2017photo, dosovitskiy2016generating}, defined on the VGG-19 feature space. The perceptual loss is formulated as
\begin{equation}
    \ell_{perc} = \sum_{l \in L}\frac{1}{C_{l}H_{l}W_{l}} \left\| \Psi_{l}(I_{R}) - \Psi_{l}(I_{GT}) \right\| ^{2}
\end{equation}
where $\Psi_{l}$ represents the features from the $l$-th layer of a pre-trained VGG-19 model and $L = \{ \texttt{relu\_2\_2}, \texttt{relu\_3\_4},\\ \texttt{relu\_4\_4}, \texttt{conv\_5\_4} \}$. We also experimented using VGG-Face \cite{parkhi2015deep} for the perceptual loss, in particular by extracting the output taken from the third convolutional layer of the fifth block before the ReLU activation, but the results were worse than with VGG-19.

The photo-realistic loss also contains two terms. First, we used the style loss \cite{gatys2016image} that is defined on the Gram matrix of the feature map for each layer in $L$

\begin{equation}
\resizebox{\columnwidth}{!}{
    $\ell_{style} = \sum_{l \in L}\frac{1}{C_{l}H_{l}W_{l}} \left\| \Psi_{l}(I_{R})^{T}\Psi_{l}(I_{R}) - \Psi_{l}(I_{GT})^{T}\Psi_{l}(I_{GT}) \right\| ^{2}$
}
\end{equation}

Second, we employed the hinge version of the adversarial loss \cite{lim2017geometric, zhang2019self}. We adopted multi-scale discriminators \cite{wang2018high}, that is 4 discriminators that have the same network structure but operate at different image scales. The adversarial loss can be formulated as
\begin{align}\label{eq:adv_loss}
\begin{split}
\nonumber   \ell_{adv, D} = - \sum_{r \in R} \bigg[   \mathbb{E}_{I^{{\downarrow r}}_{GT} \sim P(I^{{\downarrow r}}_{GT})} \Big[  min\Big(0, -1 + D(I^{{\downarrow r}}_{GT})\Big) \Big] +  \\ 
    \mathbb{E}_{I^{{\downarrow r}}_{R} \sim P(I^{{\downarrow r}}_{R})} \Big[ min\Big(0, -1 - D(I^{{\downarrow r}}_{R})\Big) \Big] \bigg]
\end{split} \\
\begin{split}
\ell_{adv, G} = - \sum_{r \in R} \lambda_{adv, r} \; \mathbb{E}_{I^{{\downarrow r}}_{D} \sim P(I^{{\downarrow r}}_{D})} \Big[  D\Big(G(I^{{\downarrow r}}_{D})\Big) \Big]
\end{split}
\end{align}
where $_{\downarrow r}$ denotes the downsampling operation with scale factor $r \in R = \{1,2,4,8\}$ and $\lambda_{adv,r}$ are the trade-off parameters for each scale discriminator. 
$\ell_{adv, D}$ and $\ell_{adv, G}$ are used to update respectively the discriminators and the generator. To stabilize the learning of the discriminators we adopted SNGAN \cite{miyato2018spectral}, incorporating the spectral normalization after each convolutional layer of the discriminator. Spectral normalization is based on regularizing the spectral norm of each layer of the discriminator by simply dividing the weight matrix by its largest eigenvalue.

The overall training loss is defined as
\begin{equation}
\resizebox{\columnwidth}{!}{%
    $\ell_{total} = \lambda_{MSE}\ell_{MSE} + \lambda_{perc}\ell_{perc} + \lambda_{style}\ell_{style} + \lambda_{adv}\ell_{adv, G}$
    }
\end{equation}
where $\lambda_{MSE}$, $\lambda_{perc}$, $\lambda_{style}$, and $\lambda_{adv}$ are the tradeoff parameters.

%% file: sec/4_results.tex
\section{Experimental Results}\label{sec:results}

\subsection{Datasets}
Similarly to \cite{galteri2020increasing}, we used the Deep Fake Detection (DFD) dataset \cite{rossler2019faceforensics}, which is composed of 363 high-resolution and high-quality videos depicting different activities performed by 28 actors. Then, we selected 55 videos of actions in which the actor is talking while facing the camera as in a setup of a video conference (\ie ``podium speech" and “talking against wall" scenes) for an overall size of $\sim$ 40 GB and a duration of $\sim$ 40 minutes. The first 22 identities were utilized for training and the last 6 for testing.

We also employed the High-Definition Talking Face (HDTF) dataset \cite{zhang2021flow}, which contains 362 videos collected from YouTube with a resolution of 720P or 1080P. We used the ``WDA" subset since it is composed of the videos that have the highest quality among those in the whole dataset, for a total of 193 videos. Since the videos have a much larger duration than those of the DFD dataset, we used only the initial 30 seconds to reduce the computational cost; this does not hamper the evaluation since the visual content remains extremely similar. We relied on this dataset only for testing purposes, to compare the proposed approach with competing state-of-the-art methods, and to evaluate the generalization capabilities of the models trained on the DFD dataset.

Starting from the raw (Constant Rate Factor 0) version of the original sequences, each video was compressed with the H.264 codec and CRF 32 and 42 using \texttt{FFmpeg} \cite{ffmpeg}. Then, only during training, the frames of each sequence were extracted by sampling one frame every five, both for the raw and compressed versions. In addition, for the compressed versions, the frames were extracted starting from a given offset measured in the number of frames to skip. This was because for the training the reference frames (\ie the raw ones) need to precede the compressed ones. The offset used in the experiments was equal to 5.

Both for training and testing we relied on \texttt{dlib} \cite{dlib09} to detect the face rectangle and the 68 facial landmarks of each frame. Then, we leveraged an affine transformation to perform the crop and alignment of the detected faces based on the set of facial landmarks. Each reference image was warped to the corresponding degraded one with Moving Least Squares to reduce the difference in pose and expression. To this end, we extracted the facial landmarks of both images and then applied the MLS algorithm presented in \cref{sec:MLS}. Finally, we used the facial landmarks of the compressed frame to generate the landmarks binary images. After the preprocessing, we ended up with 9,007 images for the training set and 12,568 images for the test set, considering the DFD dataset. Instead, all the 175,832 frames of the HDTF dataset were used for testing.

\subsection{Training Setup}
To train both the generator and the discriminator we employed the ADAM optimizer \cite{kingma2014adam} with batch size 4, learning rate $10^{-4}$ and momentum parameters $\beta_{1}=0.9$ and $\beta_{2}=0.99$. We trained all the models for 15 epochs because after that the outputs did not change significantly. We adopted several data augmentation techniques, such as shifting, $90^{\circ}$ rotations and cutout \cite{devries2017improved}. We performed a grid search to find the optimal trade-off parameters for the training losses. After that, they were set as follows: $\lambda_{MSE} = 300$, $\lambda_{perc} = 10$, $\lambda_{style} = 1$, $\lambda_{adv} = 2$, $\lambda_{adv,1}=4$, $\lambda_{adv,2}=2$, $\lambda_{adv,4}=1$ and $\lambda_{adv,8}=1$. The 4 layers used to compute the perceptual loss were given the same weight, equal to 1. During testing, we set the maximum cardinality of the set of keyframes to 10.

\subsection{Evaluation Metrics}
The performance is evaluated using six full-reference and two no-reference visual quality metrics. Regarding the full-reference metrics, we employed: \textit{1)} Peak Signal-to-Noise Ratio (PSNR), which is often used to evaluate reconstruction and compression artifacts reduction, despite its issues in estimating the perceived quality \cite{huynh2008scope, huynh2012accuracy}; \textit{2)} Structural Similarity Index Measure (SSIM) \cite{wang2004image}, another commonly used metric, although it is known that it doesn't perform well on the output of generative models \cite{ko-2020}; \textit{3)} Learned Perceptual Image Patch Similarity (LPIPS) \cite{zhang2018unreasonable}, using, in particular, the version with AlexNet \cite{krizhevsky2012imagenet} backbone. Typically LPIPS measures are in contrast with SSIM, \ie distortions that are low for LPIPS are high in SSIM and vice-versa. LPIPS has been shown to have a very strong correlation with perceived visual quality; \textit{4)} CONTRastive Image QUality Evaluator-Full Reference (CONTRIQUE-FR) \cite{madhusudana2022image}, using, in particular, the LIVE$\_$FR model downloaded from the official repository;
\textit{5)} Video Multimethod Assessment Fusion (VMAF) \cite{li2016toward}, a full reference perceptual video quality assessment model that combines multiple elementary quality metrics; \textit{5)} Video Multimethod Assessment Fusion - No Enhancement Gain (VMAF-NEG) \cite{li2020toward}, which subtracts the effect of image enhancement from the VMAF score. Indeed, VMAF tends to overpredict the perceptual quality when image enhancement techniques, such as sharpening or histogram equalization, are performed \cite{li2020toward}. Both VMAF and VMAF-NEG include an elementary metric that accounts for the temporal difference between adjacent frames of the videos, thus evaluating the presence of motion jitter and flicker.
Regarding the no-reference metrics, we relied on:  \textit{1)} Blind/Referenceless Image Spatial QUality Evaluator (BRISQUE) \cite{mittal2011blind}, which evaluates the naturalness of an image; \textit{2)} CONTRastive Image QUality Evaluator (CONTRIQUE) \cite{madhusudana2022image}, using, in particular, the LIVE model downloaded from the official repository. 

\hide{In the first sets of experiments we evaluate the impact of the keyframe set update policy, in particular: \textit{i)} how the size of the set of keyframes selected for the reconstruction, and \textit{ii)} the update policy used to maintain it. \hide{, and \textit{iii)} the source of the keyframes has an impact on the performance of the reconstruction.}
In the second set of experiments, we compare the proposed method with competing state-of-the-art approaches on the two datasets using \textit{i)} objective visual quality metrics and \textit{ii)} a user study.}

\subsection{Baselines}
We compare the proposed approach with several state-of-the-art methods: six methods for blind face restoration, HiFaceGAN \cite{yang2020hifacegan}, PSFR-GAN \cite{chen2020progressive}, GFP-GAN \cite{wang2021gfpgan}, GPEN \cite{yang2021GPEN}, DFDNet \cite{li2020blind} and ASFFNet \cite{li2020enhanced}, and one for face super-resolution, GWAINet \cite{dogan2019exemplar}. DFDNet, PSFR-GAN, GPEN and GFP-GAN do not use a reference image but utilize extra face prior, respectively some offline-generated dictionaries of facial components, a segmentation mask, and pre-trained GANs. Instead, GWAINet exploits a reference image that is warped to the compressed one by means of a warper network. HiFaceGAN does not require any additional information w.r.t.~the compressed input image. The most similar to our work is ASFFNet, which leverages a reference image and a binary landmark image. As ASFFNet needs a given static set of reference images, we make all the keyframes in the video available to it as possible guidance. Therefore, ASFFNet actually has an advantage over our approach, as, in our case, we limit the maximum cardinality of the set of keyframes to 10.

\subsection{Quantitative Results} \label{sec:quantitative_results}
The quantitative results for the DFD dataset are reported in \cref{tab:DFD_baselines_comparison}. The proposed method achieves the best performance for the LPIPS metric, which is the most indicative full-reference perceptual metric, as well as in terms of CONTRIQUE, CONTRIQUE-FR, and VMAF-NEG. PSFR-GAN performs better with regard to the signal metrics PSNR and SSIM, while GWAINet achieves the best result for BRISQUE. However, manual inspection shows that the images produced by GWAINet include excessive high-frequency artifacts and thus we did not consider this approach in the other experiments. GFP-GAN obtains the best VMAF value, probably because of its tendency to saturate colors and increase contrast at the cost of loss of photorealism, as is visible from the qualitative results. This tendency is similar to the application of image enhancement methods, which are known to boost the VMAF score \cite{li2020toward}. In support of this theory, we can notice the large difference from the VMAF-NEG score, which in contrast is not affected by image enhancement techniques. Our method achieves both the second-best VMAF value and the best VMAG-NEG value, proving its ability to obtain great overall video quality while preserving photorealism. Moreover, the VMAF and VMAF-NEG scores show that our video results are temporal consistent and do not present too much motion jitter and flicker or mosquito noise.

In the second experiment, reported in \cref{tab:HDTF_baselines_comparison}, we compare the proposed method with the baselines on the HDTF dataset. It is important to note that our model has not been trained on this dataset so that we can evaluate its generalization capabilities. Again, the proposed approach outperforms the other methods in terms of LPIPS, CONTRIQUE, CONTRIQUE-FR, and VMAF-NEG. Manual examination of the results shows that this may be motivated by the fact that several competing approaches tend to add (or, on the opposite, hide) skin imperfections or boost excessively the color of lips and eyebrows.

Overall, our method is the one that performs best with the highest consistency, as none of the baselines achieves better performance on multiple metrics simultaneously. The results obtained for the HDTF dataset also prove that the proposed model is capable of generalization. In addition, we argue that the metrics for which our method performs best, namely LPIPS, CONTRIQUE, CONTRIQUE-FR, and VMAF-NEG are those that correlate best with the actual quality of the restored frames. In \cref{app:quantitative_results_analysis} we provide some examples that support this argument.

\begin{table*}
\centering
\caption{Quantitative comparison between the proposed approach and other state-of-the-art methods for CRF 42 on DFD dataset \cite{rossler2019faceforensics}. Best and second best results are in bold and underlined, respectively. $\uparrow =$ higher values are better, $\downarrow =$ lower values are better.}
\label{tab:DFD_baselines_comparison}
\resizebox{\linewidth}{!}{%
\begin{tabular}{lcccccccc}
\toprule
Method & PSNR $\uparrow$ & SSIM $\uparrow$ & LPIPS $\downarrow$ & BRISQUE $\downarrow$ & CONTRIQUE $\downarrow$ & CONTRIQUE-FR $\downarrow$ & VMAF $\uparrow$ & VMAF-NEG $\uparrow$ \\ \midrule
GWAINet \cite{dogan2019exemplar} & 22.25 & 0.608 & 0.129 & \textbf{24.18} & 50.16 & 20.79 & 44.65 & 36.60 \\ 
HiFaceGAN \cite{yang2020hifacegan} & \underline{29.38} & 0.828 & 0.075 & 28.41 & 48.75 & 18.67 & 47.77 & 45.11 \\
PSFR-GAN \cite{chen2020progressive} & \textbf{29.68} & \underline{0.833} & \underline{0.057} & 29.07 & 46.87 & 16.46 & 48.55 & 46.22 \\ 
GFP-GAN \cite{wang2021gfpgan} & 27.51 & 0.822 & 0.081 & 34.17 & 50.84 & 23.01 & \textbf{57.55} & 48.51 \\ 
GPEN \cite{yang2021GPEN} & 27.61 & 0.813 & 0.075 & 28.67 & 49.42 & 21.36 & 55.86 & \underline{49.26} \\
DFDNet \cite{li2020blind} & 27.03 & 0.827 & 0.065 & 32.38 & 46.84 & \underline{16.04} & 55.15 & 48.95 \\
ASFFNet \cite{li2020enhanced} & 28.29 & \textbf{0.834} & 0.062 & 29.67 & \underline{46.27} & 17.48 & 51.74 & 46.84 \\ \midrule \rowcolor{tabhighlight}
\textbf{Ours} & 26.19 & 0.779 & \textbf{0.037} & \underline{27.41} & \textbf{44.95} & \textbf{13.16} & \underline{56.87} & \textbf{54.20} \\ \bottomrule
\end{tabular}%
}
\end{table*}

\begin{table*}
\centering
\caption{Quantitative comparison between the proposed approach and other state-of-the-art methods for CRF 42 on HDTF dataset \cite{zhang2021flow}. Best and second best results are in bold and underlined, respectively. $\uparrow =$ higher values are better, $\downarrow =$ lower values are better.}
\label{tab:HDTF_baselines_comparison}
\resizebox{\linewidth}{!}{%
\begin{tabular}{lcccccccc}
\toprule
Method & PSNR $\uparrow$ & SSIM $\uparrow$ & LPIPS $\downarrow$ & BRISQUE $\downarrow$ & CONTRIQUE $\downarrow$ & CONTRIQUE-FR $\downarrow$ & VMAF $\uparrow$ & VMAF-NEG $\uparrow$ \\ \midrule
HiFaceGAN \cite{yang2020hifacegan} & \textbf{30.70} & \textbf{0.864} & 0.047 & 31.71 & \underline{41.50} & \underline{10.69} & 44.50 & 42.40 \\ 
PSFR-GAN \cite{chen2020progressive} & 30.31 & 0.853 & \underline{0.046} & \textbf{30.01} & 44.99 & 13.57 & 40.18 & 38.40 \\ 
GFP-GAN \cite{wang2021gfpgan} & 28.19 & 0.846 & 0.064 & 35.02 & 46.80 & 13.95 & \textbf{51.06} & 41.27 \\ 
GPEN \cite{yang2021GPEN} & 27.72 & 0.817 & 0.061 & \underline{30.62} & 46.15 & 13.31 & 47.52 & 40.57 \\
DFDNet \cite{li2020blind} & 28.16 & 0.847 & 0.050 & 32.10 & 47.50 & 12.04 & \underline{49.48} & \underline{43.90} \\
ASFFNet \cite{li2020enhanced} & 27.88 & 0.835 & 0.058 & 32.18 & 45.39 & 12.90 & 40.15 & 35.72 \\ \midrule \rowcolor{tabhighlight}
\textbf{Ours} & \underline{30.39} & \underline{0.862} & \textbf{0.028} & 33.34 & \textbf{37.35} & \textbf{7.76} & 47.82 & \textbf{45.07} \\ \bottomrule
\end{tabular}%
}
\end{table*}

\subsection{Qualitative Results} \label{sec:qualitative_results}
\begin{figure*}
    \captionsetup[subfigure]{labelformat=empty, font=small}
    \centering
    
    \includegraphics[width=0.1\linewidth]{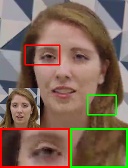}%
    \includegraphics[width=0.1\linewidth]{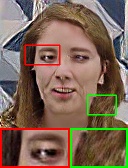}%
    \includegraphics[width=0.1\linewidth]{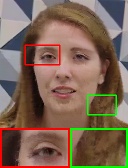}%
    \includegraphics[width=0.1\linewidth]{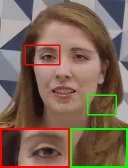}%
    \includegraphics[width=0.1\linewidth]{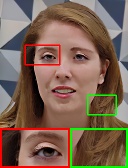}%
    \includegraphics[width=0.1\linewidth]{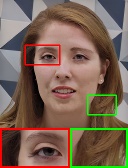}%
    \includegraphics[width=0.1\linewidth]{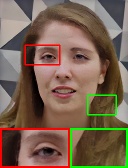}%
    \includegraphics[width=0.1\linewidth]{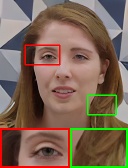}%
    \includegraphics[width=0.1\linewidth]{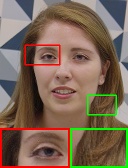}%
    \includegraphics[width=0.1\linewidth]{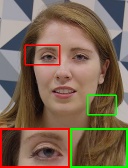}%
    
    \vspace{1pt}
    \includegraphics[width=0.1\linewidth]{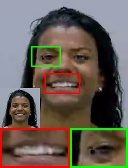}%
    \includegraphics[width=0.1\linewidth]{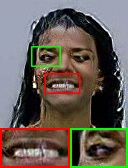}%
    \includegraphics[width=0.1\linewidth]{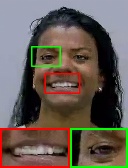}%
    \includegraphics[width=0.1\linewidth]{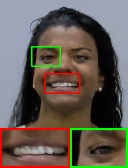}%
    \includegraphics[width=0.1\linewidth]{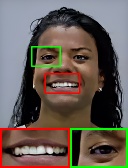}%
    \includegraphics[width=0.1\linewidth]{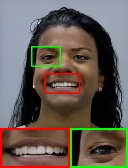}%
    \includegraphics[width=0.1\linewidth]{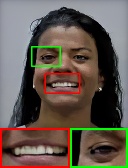}%
    \includegraphics[width=0.1\linewidth]{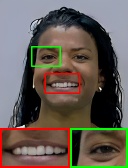}%
    \includegraphics[width=0.1\linewidth]{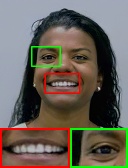}%
    \includegraphics[width=0.1\linewidth]{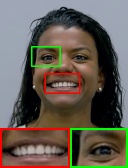}%

    \vspace{1pt}
    \includegraphics[width=0.1\linewidth]{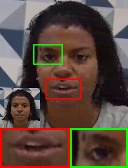}%
    \includegraphics[width=0.1\linewidth]{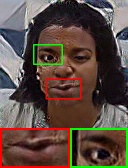}%
    \includegraphics[width=0.1\linewidth]{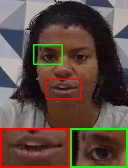}%
    \includegraphics[width=0.1\linewidth]{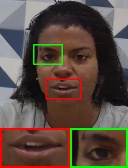}%
    \includegraphics[width=0.1\linewidth]{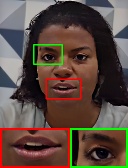}%
    \includegraphics[width=0.1\linewidth]{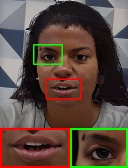}%
    \includegraphics[width=0.1\linewidth]{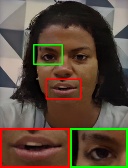}%
    \includegraphics[width=0.1\linewidth]{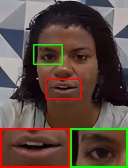}%
    \includegraphics[width=0.1\linewidth]{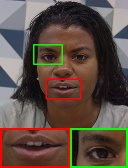}%
    \includegraphics[width=0.1\linewidth]{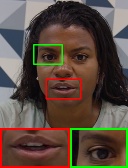}%

    \vspace{1pt}
    \includegraphics[width=0.1\linewidth]{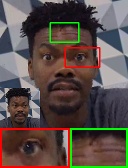}%
    \includegraphics[width=0.1\linewidth]{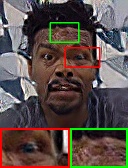}%
    \includegraphics[width=0.1\linewidth]{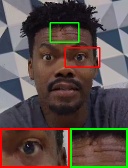}%
    \includegraphics[width=0.1\linewidth]{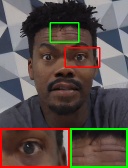}%
    \includegraphics[width=0.1\linewidth]{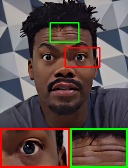}%
    \includegraphics[width=0.1\linewidth]{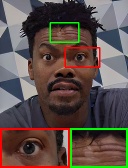}%
    \includegraphics[width=0.1\linewidth]{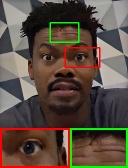}%
    \includegraphics[width=0.1\linewidth]
    {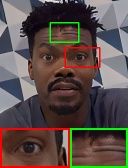}%
    \includegraphics[width=0.1\linewidth]{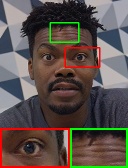}%
    \includegraphics[width=0.1\linewidth]{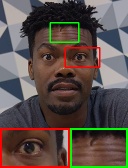}%

    \vspace{1pt}
    \includegraphics[width=0.1\linewidth]{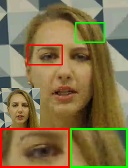}%
    \includegraphics[width=0.1\linewidth]{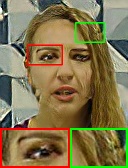}%
    \includegraphics[width=0.1\linewidth]{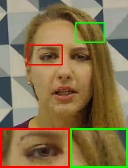}%
    \includegraphics[width=0.1\linewidth]{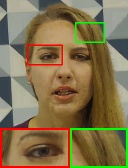}%
    \includegraphics[width=0.1\linewidth]{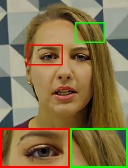}%
    \includegraphics[width=0.1\linewidth]{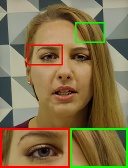}%
    \includegraphics[width=0.1\linewidth]{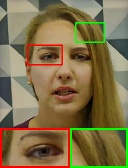}%
    \includegraphics[width=0.1\linewidth]{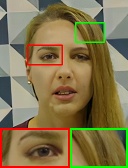}%
    \includegraphics[width=0.1\linewidth]{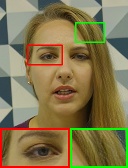}%
    \includegraphics[width=0.1\linewidth]{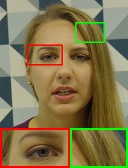}%

    \vspace{-9pt}
    \subfloat[][Input]{\includegraphics[width=0.1\linewidth]{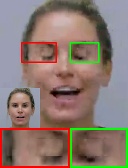}}%
    \subfloat[][GWAINet]{\includegraphics[width=0.1\linewidth]{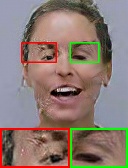}}%
    \subfloat[][HiFaceGAN]{\includegraphics[width=0.1\linewidth]{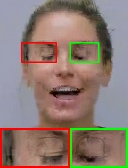}}%
    \subfloat[][PSFR-GAN]{\includegraphics[width=0.1\linewidth]{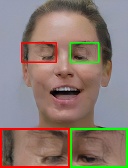}}%
    \subfloat[][GFP-GAN]{\includegraphics[width=0.1\linewidth]{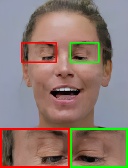}}%
    \subfloat[][GPEN]{\includegraphics[width=0.1\linewidth]{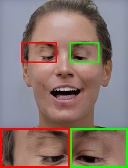}}%
    \subfloat[][DFDNet]{\includegraphics[width=0.1\linewidth]{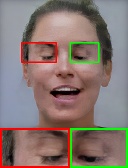}}%
    \subfloat[][ASFFNet]{\includegraphics[width=0.1\linewidth]{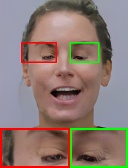}}%
    \subfloat[][\textbf{Ours}]{\includegraphics[width=0.1\linewidth]{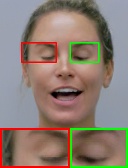}}%
    \subfloat[][Ground truth]{\includegraphics[width=0.1\linewidth]{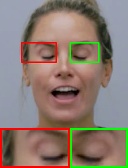}}
    \caption{Qualitative comparison between the proposed approach and the baselines for the DFD dataset and CRF 42. The bottom-left image in the input column represents the reference frame exploited by our approach. Best viewed in full screen.}
    \label{fig:baselines_comparison_dfd}
\end{figure*}

Qualitative results for the DFD dataset are shown in \cref{fig:baselines_comparison_dfd}. Our approach outperforms all the baselines in generating photorealistic and detailed results. GWAINet, HiFaceGAN and PSFR-GAN produce unsatisfactory images that still present visible artifacts, see for example the mouth in the second row. GFP-GAN and GPEN generate detailed but artsy and not photorealistic results, as the eyes in the first and fifth rows. DFDNet and ASFFNet achieve a better tradeoff between details and photorealism but, as can be seen in the last row, still produce visible artifacts. Our model exploits the reference keyframe and reproduces the high-frequency details lost after such strong compression without loss of photorealism. It is interesting to note that often the reference image (\ie the bottom-left image in the input column) is not too similar to the degraded image, but the proposed method is still able to exploit it. For example, in the last row, the reference image has open eyes while the compressed one has them closed, and despite this, our model correctly depicts the restored frame with closed eyes.

\begin{figure*}
    \captionsetup[subfigure]{labelformat=empty, font=small}
    \centering

    \includegraphics[width=0.111111\linewidth]{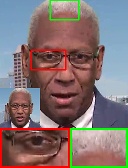}%
    \includegraphics[width=0.111111\linewidth]{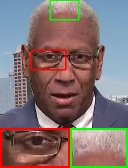}%
    \includegraphics[width=0.111111\linewidth]{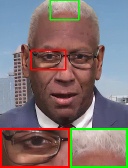}%
    \includegraphics[width=0.111111\linewidth]{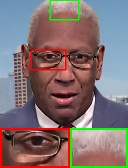}%
    \includegraphics[width=0.111111\linewidth]{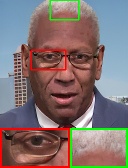}%
    \includegraphics[width=0.111111\linewidth]{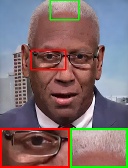}%
    \includegraphics[width=0.111111\linewidth]{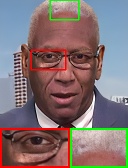}%
    \includegraphics[width=0.111111\linewidth]{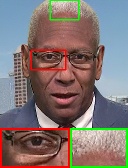}%
    \includegraphics[width=0.111111\linewidth]{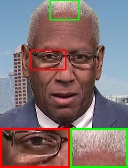}%

    \vspace{1pt}
    \includegraphics[width=0.111111\linewidth]{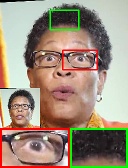}%
    \includegraphics[width=0.111111\linewidth]{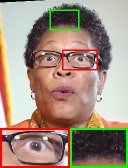}%
    \includegraphics[width=0.111111\linewidth]{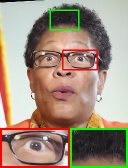}%
    \includegraphics[width=0.111111\linewidth]{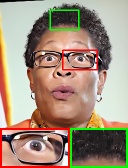}%
    \includegraphics[width=0.111111\linewidth]{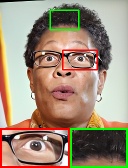}%
    \includegraphics[width=0.111111\linewidth]{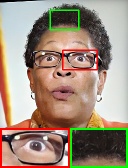}%
    \includegraphics[width=0.111111\linewidth]{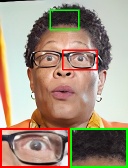}%
    \includegraphics[width=0.111111\linewidth]{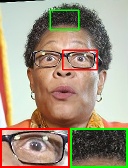}%
    \includegraphics[width=0.111111\linewidth]{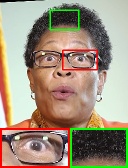}%

    \vspace{1pt}
    \includegraphics[width=0.111111\linewidth]{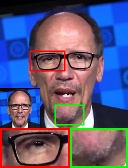}%
    \includegraphics[width=0.111111\linewidth]{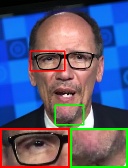}%
    \includegraphics[width=0.111111\linewidth]{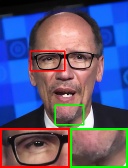}%
    \includegraphics[width=0.111111\linewidth]{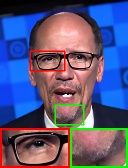}%
    \includegraphics[width=0.111111\linewidth]{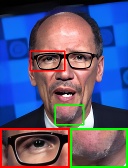}%
    \includegraphics[width=0.111111\linewidth]{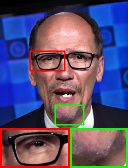}%
    \includegraphics[width=0.111111\linewidth]{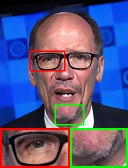}%
    \includegraphics[width=0.111111\linewidth]{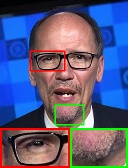}%
    \includegraphics[width=0.111111\linewidth]{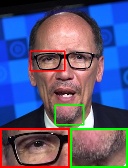}%

    \vspace{1pt}
    \includegraphics[width=0.111111\linewidth]{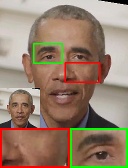}%
    \includegraphics[width=0.111111\linewidth]{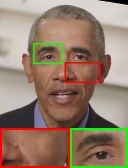}%
    \includegraphics[width=0.111111\linewidth]{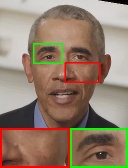}%
    \includegraphics[width=0.111111\linewidth]{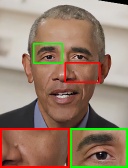}%
    \includegraphics[width=0.111111\linewidth]{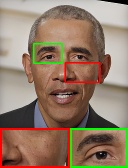}%
    \includegraphics[width=0.111111\linewidth]{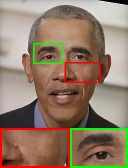}%
    \includegraphics[width=0.111111\linewidth]{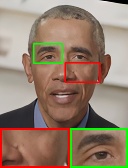}%
    \includegraphics[width=0.111111\linewidth]{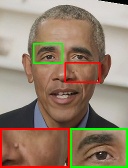}%
    \includegraphics[width=0.111111\linewidth]{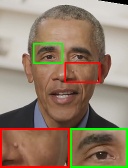}%

    \vspace{1pt}
    \includegraphics[width=0.111111\linewidth]{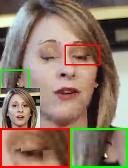}%
    \includegraphics[width=0.111111\linewidth]{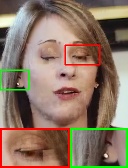}%
    \includegraphics[width=0.111111\linewidth]{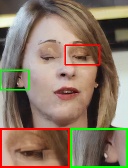}%
    \includegraphics[width=0.111111\linewidth]{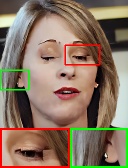}%
    \includegraphics[width=0.111111\linewidth]{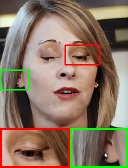}%
    \includegraphics[width=0.111111\linewidth]{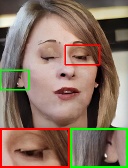}%
    \includegraphics[width=0.111111\linewidth]{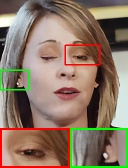}%
    \includegraphics[width=0.111111\linewidth]{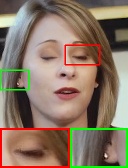}%
    \includegraphics[width=0.111111\linewidth]{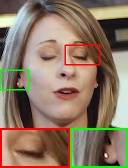}%

    \vspace{-9pt}
    \subfloat[][Input]{\includegraphics[width=0.111111\linewidth]{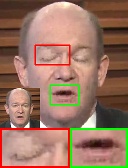}}%
    \subfloat[][HiFaceGAN]{\includegraphics[width=0.111111\linewidth]{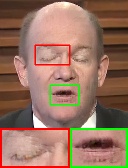}}%
    \subfloat[][PSFR-GAN]{\includegraphics[width=0.111111\linewidth]{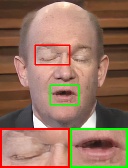}}%
    \subfloat[][GFP-GAN]{\includegraphics[width=0.111111\linewidth]{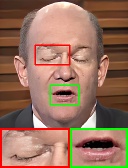}}%
    \subfloat[][GPEN]{\includegraphics[width=0.111111\linewidth]{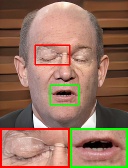}}%
    \subfloat[][DFDNet]{\includegraphics[width=0.111111\linewidth]{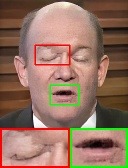}}%
    \subfloat[][ASFFNet]{\includegraphics[width=0.111111\linewidth]{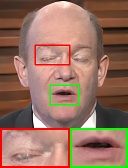}}%
    \subfloat[][\textbf{Ours}]{\includegraphics[width=0.111111\linewidth]{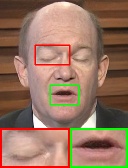}}%
    \subfloat[][GT]{\includegraphics[width=0.111111\linewidth]{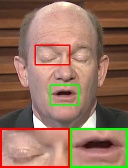}}
    \caption{Qualitative comparison between the proposed approach and the baselines for the HDTF dataset and CRF 42. The bottom-left image in the input column represents the reference frame exploited by our approach. Best viewed in full screen.}
    \label{fig:baselines_comparison_hdtf}
\end{figure*}

\Cref{fig:baselines_comparison_hdtf} shows the qualitative results for the HDTF dataset. Again, our method produces the most detailed and photorealistic images. All the baselines generate blurry hair in both the first and second rows, as well as a not detailed beard in the third row. In the first row, PSFR-GAN, GFP-GAN, GPEN and ASFFNet mistake the shadow of the glasses for their border and thus produce unrealistic results. In the fourth row, GPEN and DFDNet hallucinate moles that are not present in the ground truth. In the fifth row, our method is the only one capable of depicting the eyes as closed without adding artifacts. In the last row, GFP-GAN and GPEN add traces of glasses, while ASFFNet exploits the reference incorrectly and portrays the eyes as open. In general, our method is the one that most consistently generates satisfactory results that are similar to the ground truth.

\subsection{Subjective Experiments}
\begin{figure}
    \centering
    \includegraphics[width=0.93\columnwidth]{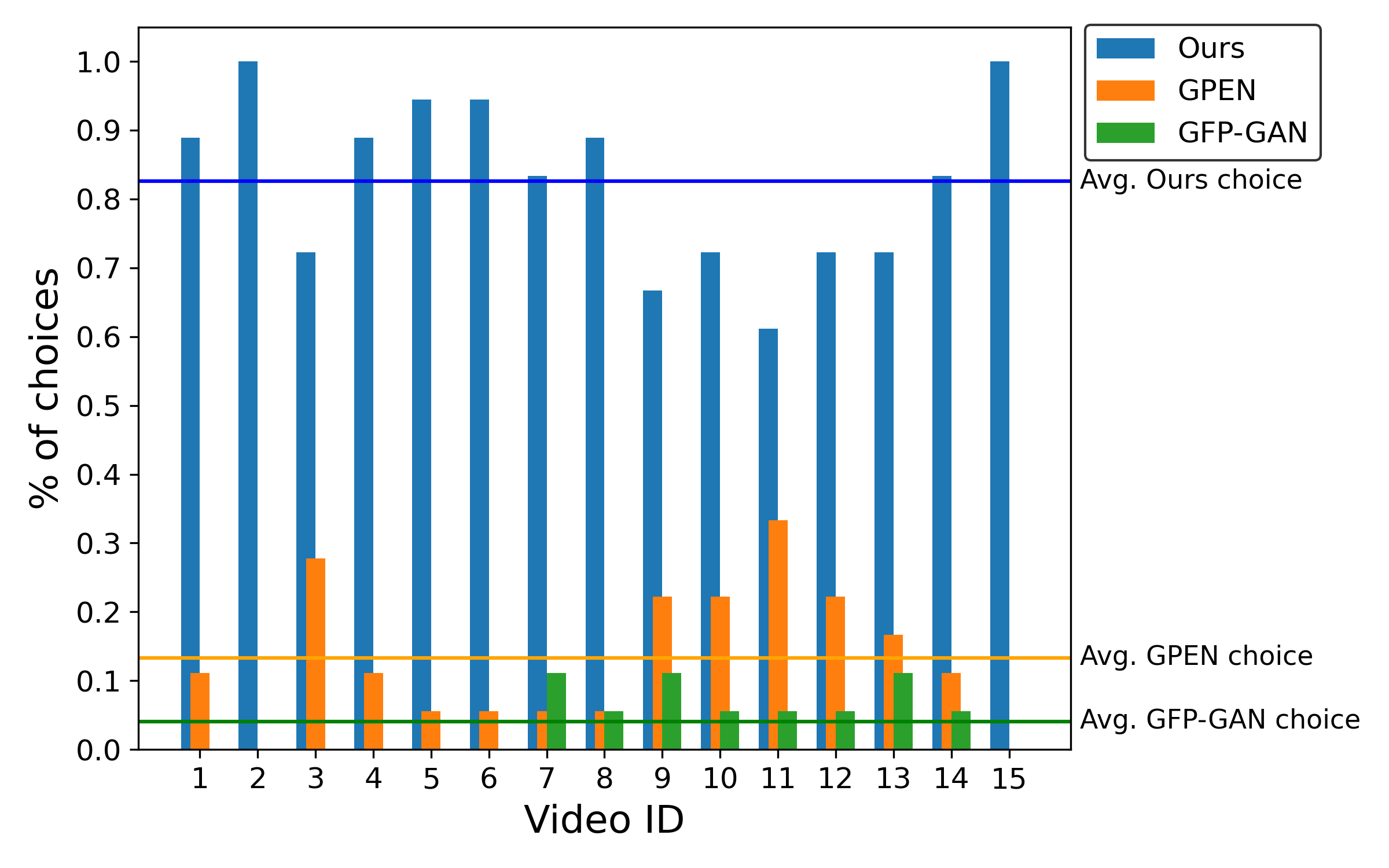}
    \vspace{-2ex}
    \caption{Subjective results using 3-AFC. Videos from 1 to 8 belong to the DFD dataset, the others to the HDTF dataset.}
    \label{fig:subjective_results}
\end{figure}

In this experiment we conducted a subjective test based on the three-alternative forced choice (3-AFC) methodology, using the \textit{AVrate Voyager} tool \cite{rao2021crowd, goering2021voyager}. The test included the inspection of 15 sets of videos, 8 from the DFD dataset and 7 from the HDTF one, so as to maintain the completion time of around 15-20 minutes and avoid excessive fatigue as recommended by ITU-R BT.500-13 \cite{ITU-R-2012}. Each original video was compressed with CRF 42 and restored using our proposed method, GPEN \cite{yang2021GPEN} and GFP-GAN \cite{wang2021gfpgan}; using 3-AFC allowed to reduce the number of required comparisons \cite{ciqb-2017}. Participants (18, \ie almost double the minimum required \cite{Winkler-2009}) were requested to choose the reconstruction that matched more closely the original high-quality video, without considering aesthetic preferences. The position of the results of all the methods was changed randomly for each evaluation. \Cref{fig:subjective_results} reports the percentages of the forced choices for the 15 sets. The much larger preference given to our proposed method can be attributed to the fact the proposed GAN introduces fewer high-frequency details and color shifts than the GPEN \cite{yang2021GPEN} and GFP-GAN \cite{wang2021gfpgan}; these additions tend to be more visible in a video sequence than when evaluating separately the frames using the quality metrics.

\subsection{Inference Time}
We compared the Frames Per Second (FPS) processed by our model with the baselines. The experiments were performed on an NVIDIA RTX 2080 Ti GPU. As shown in \cref{tab:inference_time}, our method achieves a number of FPS similar to or better than the baselines but outperforms them in terms of quality. Given that our model runs at almost 45 FPS, it proves to be capable of real-time inference and therefore suitable for videoconferencing. 

\begin{table}
\centering
\caption{FPS comparison between the proposed approach and other state-of-the-art methods. Best and second best results are in bold and underlined, respectively.}\label{tab:inference_time}
\huge
\resizebox{0.8\linewidth}{!}{ 
\begin{tabular}{lcc}
\toprule
Method & \# parameters (M) & FPS \\  \midrule
HiFaceGAN \cite{yang2020hifacegan} & 72.22 & \underline{44} \\
PSFR-GAN \cite{chen2020progressive} & \underline{67.26} & 28 \\
GFP-GAN \cite{wang2021gfpgan} & 86.44 & \textbf{49} \\
GPEN \cite{yang2021GPEN} & 71.00 & 39 \\
DFDNet \cite{li2020blind} & 113.31 & 4 \\
ASFFNet \cite{li2020enhanced} & \textbf{23.62} & 24 \\
\midrule \rowcolor{tabhighlight}
\textbf{Ours} & 96.35 & \underline{44} \\ 
\bottomrule
\end{tabular}%
}
\end{table}

\subsection{Ablation Studies}
\paragraph{Architecture}
We performed ablation studies to evaluate the importance of each component of our architecture. In particular, we measure the effect of using: \textit{i)} Multi-scale features; \textit{ii)} ASFF blocks; \textit{iii)} SFT blocks. We start from a single-scale features model that considers only the features with the smallest size but with the most channels and that relies on concatenation instead of the ASFF and SFT blocks. Then, we gradually add each component: first individually and then in combination with each other. The results are reported in \cref{tab:ablations}. Our experiments show how the use of multi-scale features is the most important component of the architecture, followed by the ASFF and SFT blocks. Additionally, we substitute the ASFF and SFT blocks one at a time with SPADE \cite{park2019spade}, a spatially-adaptive denormalization block. The proposed architecture outperforms both versions that make use of SPADE, proving that ASFF and SFT blocks are more effective in our architecture.

\begin{table*}
\centering
\caption{Ablation studies with CRF 42 on the DFD dataset. SSF stands for Single-Scale Features, MSF for Multi-Scale Features. Best results are in bold. $\uparrow =$ higher values are better, $\downarrow =$ lower values are better.}\label{tab:ablations}
\resizebox{\linewidth}{!}{ 
\begin{tabular}{lccccccccccc}
\toprule
Ablation & PSNR $\uparrow$ & SSIM $\uparrow$ & LPIPS $\downarrow$ & BRISQUE $\downarrow$ & CONTRIQUE $\downarrow$ & CONTRIQUE-FR $\downarrow$ & VMAF $\uparrow$ & VMAF-NEG $\uparrow$ \\  \midrule
SSF w/o ASFF w/o SFT & 25.41 & 0.736 & 0.078 & 33.30 & 49.95 & 16.50 & 48.19 & 46.19 \\ 
SSF w/ ASFF w/ SFT & 24.99 & 0.736 & 0.079 & 30.96 & 49.97 & 15.83 & 47.90 & 45.83 \\ 
MSF w/o ASFF w/o SFT & \textbf{26.19} & 0.777 & 0.039 & \textbf{25.84} & \textbf{44.34} & 13.84 & 54.49 & 52.34 \\ 
MSF w/o ASFF w/ SFT & 26.17 & 0.776 & 0.038 & 26.21 & 45.25 & 13.87 & 55.29 & 52.99 \\
MSF w/ ASFF w/o SFT & 26.12 & 0.776 & 0.038 & 29.17 & 45.86 & 13.39 & 56.33 & 53.84 \\
MSF w/ ASFF w/ SPADE & 26.17 & 0.777 & 0.038 & 28.14 & 45.06 & 14.18 & 55.18 & 52.97 \\ 
MSF w/ SPADE w/ SFT & 26.18 & 0.776 & 0.039 & 27.57 & 45.12 & 13.78 & 55.87 & 53.39 \\ \midrule \rowcolor{tabhighlight} \textbf{Ours}
(MSF w/ ASFF w/ SFT) & \textbf{26.19} & \textbf{0.779} & \textbf{0.037} & 27.41 & 44.95 & \textbf{13.16} & \textbf{56.87} & \textbf{54.20} \\ 
\bottomrule
\end{tabular}%
}
\end{table*}

\paragraph{Keyframes Selection Policy}
\Cref{tab:strategy_comparison_crf_32,tab:strategy_comparison_crf_42} compare the proposed LFU policy update method with a different approach that maximizes the diversity of the keyframes, called ``Max distance''. The ``Max distance'' policy consists of maximizing the Euclidean distance between the facial landmarks of the frames of the set, in order to have a wide range of poses and expressions. The idea is that in this way, every future frame of the video should always have a reference in the set that is not too different. For each new keyframe, its distance to all the keyframes in the set is computed. Then, between all the possible combinations of frames, we choose the group of keyframes that maximizes the total distance, so the new keyframe is not necessarily added to the set.

\begin{table}
\centering
\caption{Ablation studies on the keyframes selection policy for the DFD dataset and CRF 32. Best results are in bold. $\uparrow =$ higher values are better, $\downarrow =$ lower values are better.}
\Huge
\resizebox{\linewidth}{!}{ 
\begin{tabular}{lcccccccc}
\toprule
Strategy & PSNR $\uparrow$ & SSIM $\uparrow$ & LPIPS $\downarrow$ & BRISQUE $\downarrow$ & CONTRIQUE $\downarrow$ & CONTRIQUE-FR $\downarrow$ & VMAF $\uparrow$ & VMAF-NEG $\uparrow$ \\  
\midrule
Max distance & 29.29 & 0.844 & 0.022 & \textbf{27.89} & 43.83 & 13.79 & 66.99 & 63.53 \\ 
LFU & \textbf{29.32} & \textbf{0.845} & \textbf{0.021} & 27.92 & \textbf{43.77} & \textbf{13.75} & \textbf{67.18} & \textbf{63.69} \\
\bottomrule
\end{tabular}
} 
\label{tab:strategy_comparison_crf_32}
\end{table}

\begin{table}
\centering
\caption{Ablation studies on the keyframes selection policy for the DFD dataset and CRF 42. Best results are in bold. $\uparrow =$ higher values are better, $\downarrow =$ lower values are better.}
\Huge
\resizebox{\linewidth}{!}{ 
\begin{tabular}{lcccccccc}
\toprule
Strategy & PSNR $\uparrow$ & SSIM $\uparrow$ & LPIPS $\downarrow$ & BRISQUE $\downarrow$ & CONTRIQUE $\downarrow$ & CONTRIQUE-FR $\downarrow$ & VMAF $\uparrow$ & VMAF-NEG $\uparrow$ \\ 
\midrule
Max distance & \textbf{26.22} & 0.776 & 0.039 & 27.50 & 45.09 & 13.29 & 55.85 & 53.24 \\ 
LFU & 26.19 & \textbf{0.779} & \textbf{0.037} & \textbf{27.41} & \textbf{44.95} & \textbf{13.16} & \textbf{56.87} & \textbf{54.20} \\ 
\bottomrule
\end{tabular}
} 
\label{tab:strategy_comparison_crf_42}
\end{table}

\Cref{tab:strategy_comparison_crf_32} reports the results obtained with CRF 32 and \Cref{tab:strategy_comparison_crf_42} those for CRF 42. The maximum number of keyframes in the group was set to 10 in both cases. The results show that the proposed LFU strategy outperforms the ``Max distance'' one for almost all the metrics.

\paragraph{Keyframes Set Cardinality}
Regarding the dimension of the set of keyframes, we expect that as the maximum cardinality increases, the results will improve. In fact, having more possible references available, it is less likely that a compressed frame has no similar reference. The results reported in \cref{tab:max_num_references} confirm our assumption, but the increase in performance is not too significant. However, we set the maximum cardinality to 10 because the time needed to choose the best keyframe is still about 0.1 milliseconds so a higher number of keyframes to choose from does not impact the computational complexity significantly.

\begin{table}
\centering
\caption{Ablation studies on the maximum cardinality of the set of references for the DFD dataset and CRF 42. Best results are in bold. $\uparrow =$ higher values are better, $\downarrow =$ lower values are better.}\label{tab:max_num_references}
\Huge
\resizebox{\linewidth}{!}{ 
\begin{tabular}{ccccccccc}
\toprule
Max cardinality & PSNR $\uparrow$ & SSIM $\uparrow$ & LPIPS $\downarrow$ & BRISQUE $\downarrow$ & CONTRIQUE $\downarrow$ & CONTRIQUE-FR $\downarrow$ & VMAF $\uparrow$ & VMAF-NEG $\uparrow$ \\  \midrule
1 & 26.23 & \textbf{0.779} & 0.038 & 27.68 & 45.07 & 13.18 & 56.63 & 53.95 \\ 
3 & \textbf{26.25} & \textbf{0.779} & \textbf{0.037} & 27.50 & 45.01 & 13.19 & 56.82 & 54.14 \\ 
5 & \textbf{26.25} & \textbf{0.779} & \textbf{0.037} & 27.46 & 45.00 & \textbf{13.16} & \textbf{56.87} & \textbf{54.20} \\ 
10 & 26.19 & \textbf{0.779} & \textbf{0.037} & \textbf{27.41} & \textbf{44.95} & \textbf{13.16} & \textbf{56.87} & \textbf{54.20} \\ 
\bottomrule
\end{tabular}%
}
\end{table}

\paragraph{Feature Extractor}
In this experiment, we replace the VGG-19 backbone with different feature extractors. In particular, we exploit the small and large versions of MobileNetV3 \cite{howard2019searching}, a popular and light CNN designed for mobile platforms which, from our experiments, reduces the inference time of our model by about two times. \Cref{tab:features_extractors} reports the quantitative results. As expected, the version with the VGG-19 outperforms the MobileNetV3 ones, but the number of parameters is an order of magnitude greater. However, looking at the qualitative results obtained with the MobileNetV3 as the feature extractor we noticed how they were still more than acceptable, proving how effective our approach is, and suggesting that these backbones could be used for deployment on mobile devices.

\begin{table}[]
\centering
\caption{Ablation studies on feature extractor for the DFD dataset and CRF 42. Best results are in bold. $\uparrow =$ higher values are better, $\downarrow =$ lower values are better.}\label{tab:features_extractors}
\Huge
\resizebox{\linewidth}{!}{ 
\begin{tabular}{lcccccccccccc}
\toprule
Features extractor & \# parameters & PSNR $\uparrow$ & SSIM $\uparrow$ & LPIPS $\downarrow$ & BRISQUE $\downarrow$ & CONTRIQUE $\downarrow$ & CONTRIQUE-FR $\downarrow$ & VMAF $\uparrow$ & VMAF-NEG $\uparrow$ \\  \midrule
MobileNetV3 Small & \textbf{2.72M} & 25.87 & 0.784 & 0.047 & 33.18 & 44.96 & 13.00 & 50.30 & 48.19 \\ 
MobileNetV3 Large & 8.45M & 26.11 & \textbf{0.786} & 0.045 & 31.37 & 45.94 & \textbf{11.91} & 50.26 & 48.01 \\ 
VGG-19 & 96.35M & \textbf{26.19} & 0.779 & \textbf{0.037} & \textbf{27.41} & \textbf{44.95} & 13.16 & \textbf{56.87} & \textbf{54.20} \\ 
\bottomrule
\end{tabular}%
}
\end{table}

\paragraph{Discriminator}
We substitute the multi-scale discriminators with a standard single-scale discriminator. Consequently, we also replace the adversarial loss described in \cref{eq:adv_loss} with the following one:

\begin{align}
\begin{split}
\nonumber   \ell_{adv, D} = - \mathbb{E}_{I_{GT} \sim P(I_{GT})} \left [  min(0, -1 + D(I_{GT}))\right ] -  \\ 
    \mathbb{E}_{I_{R} \sim P(I_{R})} \left [  min(0, -1 - D(I_{R}))\right ] 
\end{split} \\
\begin{split}
\ell_{adv, G} = - \mathbb{E}_{I_{D} \sim P(I_{D})} \left [  D(G(I_{D}))\right ]
\end{split}
\end{align}
$\ell_{adv, D}$ and $\ell_{adv, G}$ were used to update respectively the discriminator and the generator.

\begin{figure}
    \centering
    \subfloat[][Single-scale]{\includegraphics[width=0.33\columnwidth]{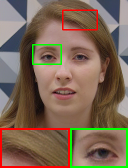}}
    \hfill
    \subfloat[][Multi-scale]{\includegraphics[width=0.33\columnwidth]{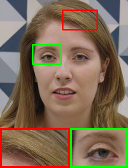}}
    \hfill
    \subfloat[][Ground truth]{\includegraphics[width=0.33\columnwidth]{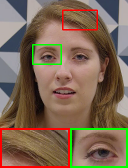}}
    \caption{Qualitative results for different discriminators for max cardinality 10 and CRF 42 on the DFD dataset.}
    \label{fig:ablation_discriminator}
\end{figure}

\Cref{tab:features_extractors_DFD,tab:features_extractors_HDTF} report the quantitative results for the DFD and HDTF datasets, respectively. Even if the version with the single-scale discriminator outperforms the multi-scale one for some metrics, the qualitative results show clearly that the use of the multi-scale discriminators allows to obtain less blurry and more sharp and detailed outputs. This is proven also by the lower values of the LPIPS metric for both datasets. For instance, \cref{fig:ablation_discriminator} shows how the multi-scale version has less blurred and more detailed hair and eyes than the single-scale one, as well as an overall color more faithful to the ground truth. Additionally, the multi-scale discriminators let to achieve higher VMAF and VMAF-NEG values, which correspond to a better temporal consistency. 

\begin{table}[]
\centering
\caption{Ablation studies on the discriminator for the DFD dataset and CRF 42. Best results are in bold. $\uparrow =$ higher values are better, $\downarrow =$ lower values are better.}\label{tab:features_extractors_DFD}
\Huge
\resizebox{\linewidth}{!}{ 
\begin{tabular}{lcccccccc}
\toprule
Discriminator & PSNR $\uparrow$ & SSIM $\uparrow$ & LPIPS $\downarrow$ & BRISQUE $\downarrow$ & CONTRIQUE $\downarrow$ & CONTRIQUE-FR $\downarrow$ & VMAF $\uparrow$ & VMAF-NEG $\uparrow$ \\  \midrule
Single-scale & \textbf{26.49} & \textbf{0.804} & 0.039 & 32.97 & \textbf{44.87} & \textbf{11.88} & 50.44 & 48.85 \\ 
Multi-scale & 26.19 & 0.779 & \textbf{0.037} & \textbf{27.41} & 44.95 & 13.16 & \textbf{56.87} & \textbf{54.20} \\ 
\bottomrule
\end{tabular}%
}
\end{table}

\begin{table}[]
\centering
\caption{Ablation studies on the discriminator for the HDTF dataset and CRF 42. Best results are in bold. $\uparrow =$ higher values are better, $\downarrow =$ lower values are better.}\label{tab:features_extractors_HDTF}
\Huge
\resizebox{\linewidth}{!}{ 
\begin{tabular}{lcccccccc}
\toprule
Discriminator & PSNR $\uparrow$ & SSIM $\uparrow$ & LPIPS $\downarrow$ & BRISQUE $\downarrow$ & CONTRIQUE $\downarrow$ & CONTRIQUE-FR $\downarrow$ & VMAF $\uparrow$ & VMAF-NEG $\uparrow$ \\  \midrule
Single-scale & 30.34 & \textbf{0.873} & 0.030 & 34.25 & 37.52 & 8.22 & 40.21 & 38.94 \\ 
Multi-scale & \textbf{30.39} & 0.862 & \textbf{0.028} & \textbf{33.34} & \textbf{37.35} & \textbf{7.76} & \textbf{47.82} & \textbf{45.07} \\ 
\bottomrule
\end{tabular}%
}
\end{table}

%% file: sec/5_conclusion.tex
\section{Conclusion}
In this paper, we have proposed a novel GAN-based method and a keyframe selection system that improves the visual quality of videoconference videos enhancing the appearance of faces. A key element of the system is the policy that updates a set of previous I-frames and exploits them to improve the visual quality improvement process. The proposed approach improves over competing state-of-the-art methods in terms of perceptual metrics and is rated much better in terms of fidelity by human evaluators.

\paragraph{Acknowledgments}
This work was partially supported by the European Commission under European Horizon 2020 Programme, grant number 101004545 - ReInHerit.

%% file: sec/X_suppl.tex
\clearpage
\maketitlesupplementary

\section*{Quantitative Results Analysis} \label{app:quantitative_results_analysis}
\begin{figure}[]
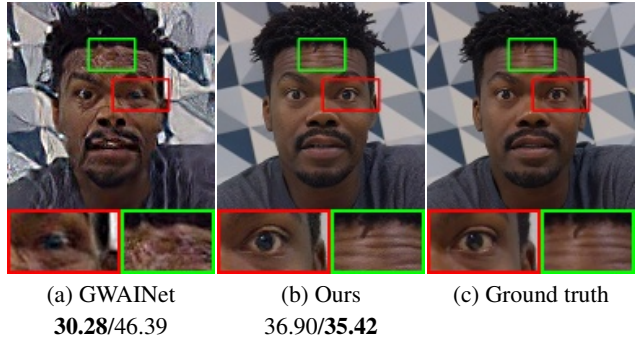

    \captionsetup[subfigure]{font=small, justification=centering}
    \centering
    \subfloat[][GWAINet \\ \vspace{1pt} \textbf{30.28}/46.39]{\includegraphics[width=0.33\columnwidth]{fig/qualitative_results/DFD/2/25_GWAIN_complete.jpg}}
    \hfill
    \subfloat[][Ours \\ \vspace{1pt} 36.90/\textbf{35.42}]{\includegraphics[width=0.33\columnwidth]{fig/qualitative_results/DFD/2/25_ours_complete.jpg}}
    \hfill
    \subfloat[][Ground truth]{\includegraphics[width=0.33\columnwidth]{fig/qualitative_results/DFD/2/25_GT_complete.jpg}}
    \caption{Comparison between our method and GWAINet \cite{dogan2019exemplar}. The reported values represent BRISQUE$\:\downarrow\,$/CONTRIQUE$\:\downarrow\,$, respectively, where $\downarrow$ means that lower values are better. Best results for each image are highlighted in bold.}
    \label{fig:app_quantitative_no_reference}
\end{figure}

\begin{figure}[]
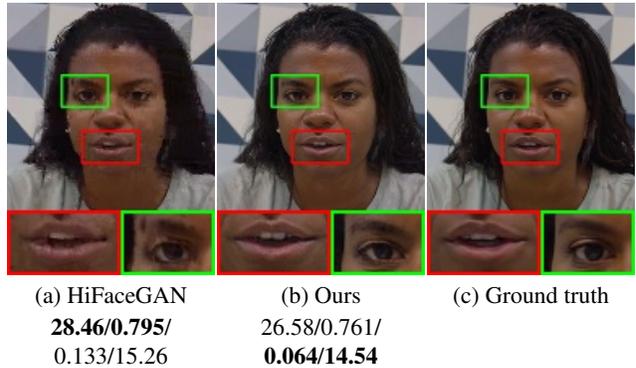

    \captionsetup[subfigure]{font=small, justification=centering}
    \centering
    \subfloat[][HiFaceGAN \\ \vspace{1pt} \textbf{28.46}/\textbf{0.795}/ \\ 0.133/15.26]{\includegraphics[width=0.33\columnwidth]{fig/qualitative_results/DFD/4/26_HiFaceGAN_complete.jpg}}
    \hfill
    \subfloat[][Ours \\ \vspace{1pt} 26.58/0.761/ \\ \textbf{0.064}/\textbf{14.54}]{\includegraphics[width=0.33\columnwidth]{fig/qualitative_results/DFD/4/26_ours_complete.jpg}}
    \hfill
    \subfloat[][Ground truth]{\includegraphics[width=0.33\columnwidth]{fig/qualitative_results/DFD/4/26_GT_complete.jpg}}
    \caption{Comparison between our approach and HiFaceGAN \cite{yang2020hifacegan}. The reported values represent PSNR$\:\uparrow\,$/SSIM$\:\uparrow\,$/LPIPS$\:\downarrow\,$/CONTRIQUE-FR$\:\downarrow\,$, respectively. $\uparrow =$ higher values are better, $\downarrow =$ lower values are better. Best results for each image are highlighted in bold.}
    \label{fig:app_quantitative_full_reference}
\end{figure}

In \cref{sec:quantitative_results} we reported the quantitative results for the DFD and HDTF datasets. Our method obtains the best performance in terms of LPIPS, CONTRIQUE, CONTRIQUE-FR and VMAF-NEG. We argue that these metrics best correlate with the perceived visual quality. In \cref{fig:app_quantitative_no_reference,fig:app_quantitative_full_reference} we show two examples supporting our argument. In \cref{fig:app_quantitative_no_reference} we compare a frame restored by our method and by GWAINet and present the corresponding values of no-reference metrics BRISQUE and CONTRIQUE. The proposed approach clearly generates a more satisfying image than GWAINet, which adds high-frequency artifacts. We argue that these artifacts deceive BRISQUE, which mistakes them for high-frequency details that are distinctive of high-quality images \cite{seidenari2022language}. In \cref{fig:app_quantitative_full_reference} we report the values of the full-reference metrics PSNR, SSIM, LPIPS and CONTRIQUE-FR obtained by our approach and HiFaceGAN for a restored frame. Again, the proposed method produces a more detailed and photorealistic image, while HiFaceGAN generates a frame with visible artifacts. However, HiFaceGAN obtains better values for PSNR and SSIM. PSNR and SSIM are signal-based metrics that do not correlate well with the perceived visual quality for the output of generative models \cite{huynh2008scope,huynh2012accuracy,ko-2020}. On the contrary, LPIPS and CONTRIQUE-FR are perceptual-based metrics and are good indicators of the actual perceived quality of an image.

\begin{figure}[]
    \captionsetup[subfigure]{font=small, justification=centering}
    \centering
    \subfloat[][GFP-GAN]{\includegraphics[width=0.33\columnwidth]{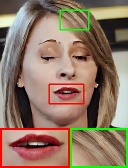}}
    \hfill
    \subfloat[][Ours]{\includegraphics[width=0.33\columnwidth]{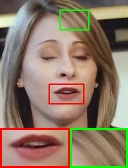}}
    \hfill
    \subfloat[][Ground truth]{\includegraphics[width=0.33\columnwidth]{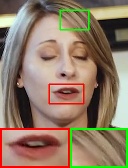}}
    \caption{Comparison between our approach and GFP-GAN.}
    \label{fig:app_quantitative_vmaf}
\end{figure}

Regarding VMAF, it is known that image enhancement techniques tend to boost its values \cite{li2020toward}. As \cref{fig:app_quantitative_vmaf} shows, some baselines, such as GFP-GAN, saturate colors and increase the contrast of the restored frames, making them more visually pleasing but less similar to the ground truth. This tendency is similar to the application of image enhancement methods. We argue that this is the reason why such baselines perform so well in terms of VMAF. Our argument is supported by the large difference between the values of the baselines for VMAF and VMAF-NEG, which is not affected by image enhancement techniques, in \cref{tab:DFD_baselines_comparison,tab:HDTF_baselines_comparison}. On the contrary, our method obtains high values for both metrics without a substantial difference between them, meaning that they are due to the actual quality and temporal consistency of the results, and not due to color enhancement.

